%% file: iclr2026_conference.tex
\documentclass{article} 
\usepackage{iclr2026_conference,times}

\input{math_commands.tex}

\usepackage{hyperref}
\usepackage{url}
\usepackage{booktabs}       
\usepackage{amsfonts}       
\usepackage{nicefrac}       
\usepackage{microtype}      
\usepackage{xcolor}         
\usepackage{multirow}
\usepackage{graphicx}
\usepackage{amsmath}  
\usepackage{algorithm}
\usepackage{algpseudocode}
\usepackage{amssymb}
\usepackage{marvosym}

\title{InterActHuman: Multi-Concept Human Animation with Layout-Aligned Audio Conditions}

\author{Zhenzhi Wang\thanks{Equal contribution.} $^1$, Jiaqi Yang$^{*2}$, Jianwen Jiang$^{*2}$\textsuperscript{\Letter}, Chao Liang$^2$, Gaojie Lin$^2$,\\  \textbf{Zerong Zheng}$^2$, \textbf{Ceyuan Yang}$^2$, \textbf{Yuan Zhang}$^2$, \textbf{Mingyuan Gao}$^2$, \textbf{Dahua Lin}$^1$ \\
$^1$The Chinese University of Hong Kong \quad $^2$ByteDance \\
}

\iclrfinalcopy 
\begin{document}

\maketitle

\begin{abstract}
End-to-end human animation with rich multi-modal conditions, e.g., text, image and audio has achieved remarkable advancements in recent years. However, most existing methods could only animate a single subject and inject conditions in a global manner, ignoring scenarios where multiple concepts could appear in the same video with rich human-human interactions and human-object interactions. Such a global assumption prevents precise and per-identity control of multiple concepts including humans and objects, therefore hinders applications. In this work, we discard the single-entity assumption and introduce a novel framework that enforces strong, region‑specific binding of conditions from modalities to each identity's spatiotemporal footprint. Given reference images of multiple concepts, our method could automatically infer layout information by leveraging a mask predictor to match appearance cues between the denoised video and each reference appearance. Furthermore, we inject local audio condition into its corresponding region to ensure layout-aligned modality matching in an iterative manner. This design enables the high-quality generation of human dialogue videos between two to three people or video customization from multiple reference images. Empirical results and ablation studies validate the effectiveness of our explicit layout control for multi-modal conditions compared to implicit counterparts and other existing methods. Video demos are available at \url{https://zhenzhiwang.github.io/interacthuman/}
\end{abstract}

\section{Introduction}
By leveraging the priors of pretrained Diffusion Transformer-based (DiT) video diffusion models~\citep{bar2024lumiere,svd,ayl,guo2024animatediff,zhou2022magicvideo,walt,wang2023modelscope,vdm,videogan,cvideogan,singer2022make,text2video,villegas2022phenaki,lin2025apt}, end-to-end human animation models, especially audio-driven approaches~\citep{he2023gaia,emo,xu2024hallo,wang2024vexpress,chen2024echomimic,xu2024vasa,stypulkowski2024diffused,jiang2024loopy,lin2024cyberhost,lin2025omnihuman} have achieved high-quality human-centric video generation and strong controllability from multi-modal conditions, such as text, image and audio. However, most existing methods commonly hold an assumption of a single-identity paradigm: all available conditions should be fused globally and implicitly assumed to describe one unique subject in the given image. Although this global injection strategy simplifies conditioning by sharing the same condition signal across all regions, it fundamentally limits scalability in scenarios involving multiple individuals or complex human-object interactions, where each entity requires distinct appearance and voice attributes.

\begin{figure}
    \centering
    \includegraphics[width=\textwidth]{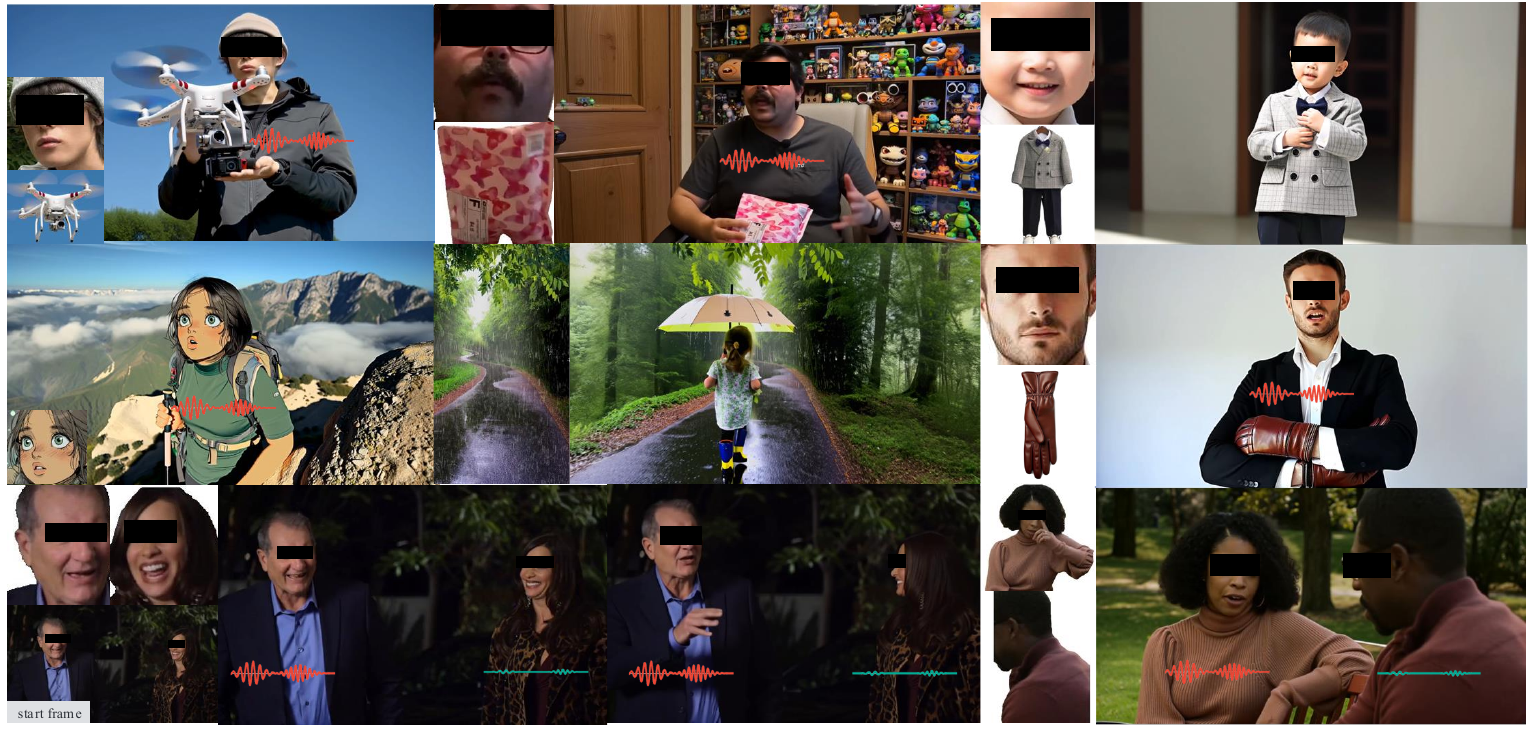}
        \vspace{-0.5em}
    \caption{Video frames generated from audio and multi-concept reference images (human heads/full bodies, objects, scenes) display rich, audio-matched expressions. Our method enables compositional generation including outfit changes, human–object interactions, anime styles, dialogues even without a start frame. Red and green wave icons denote speaking and listening, respectively.}
    \label{fig:teaser}
        \vspace{-1em}
\end{figure}

Recent multi-concept video customization methods, such as Video-Alchemist~\citep{chen2025multi}, ConceptMaster~\citep{huang2025conceptmaster}, Phantom~\citep{liu2025phantom}, and SkyReels-A2~\citep{fei2025skyreels}, enable injecting multiple reference images into a single video, facilitating multi-person or human-object interaction scenarios. However, these methods face significant challenges when directly applied to human animation tasks. In such tasks, animation signals are highly specific to individual identities and demand precise conditioning and alignment with specific spatiotemporal regions. For example, audio signals are exclusively associated with the current speaker and are unrelated to the background or other individuals' concepts. Existing customization methods, akin to end-to-end human animation approaches, still adopt video-level condition injection. While this approach may suffice for general video generation, it often causes confusion in human-centric video generation, making it difficult to produce satisfactory results, as shown in Fig.~\ref{fig:audio}.
For end-to-end human-centric video generation, conditioning inputs should ideally include not only global modalities (e.g., reference images and text descriptions) but also local modalities (e.g., audio). As mentioned, existing human animation and multi-concept customization methods fail to address this critical distinction. This limitation motivates us to propose a new framework that enables the precise injection of local human-related modalities, a capability that is both essential and urgently needed for robust multi-concept, human-centric video generation, as illustrated in Fig.~\ref{fig:teaser}.

In this paper, we propose \textbf{InterActHuman}, a video diffusion framework for spatially aligning multi-modal conditions in multi-concept human video generation. Unlike prior methods~\citep{chen2025multi,huang2025conceptmaster} that rely on feature fusion and attention to implicitly learn relationships between condition signals and concepts, InterActHuman introduces an attention module that explicitly predicts the spatial locations where the reference concepts appear in the video. This explicit layout allows the model to accurately associate local audio conditions with the correct regions via iterative mask prediction and masked audio attention during diffusion inference. Our framework offers two key advantages. First, it enforces a stronger layout constraint by precisely binding each condition to its corresponding spatial region. Second, it provides a unified interface for synchronously injecting all modalities (e.g., visual and acoustic inputs) through the layout. These features make InterActHuman well suited for multi-modal, multi-concept human animation and establish a baseline for this domain.

Although using an explicit mask for condition injection may seem straightforward, it creates a chicken-and-egg dilemma: during inference the final video is not yet available, leaving the spatial positions of each identity uncertain and making accurate mask prediction impossible; yet without those masks, spatial injection of local audio conditions cannot be performed, leading to an incomplete or misaligned generation process. To address this, we leverage the iterative denoising process inherent in diffusion models~\citep{song2021ddim}. Specifically, we introduce a mask-predictor branch into the diffusion pipeline and adopt an interleaved mask-prediction strategy, wherein the mask predicted at step $k$ guides condition injection at step $k{+}1$. This iterative refinement progressively finalizes the spatial layout, breaking the cyclic dependency and enabling precise spatial alignment even without ground-truth video during inference. In essence, our approach converts the chicken-and-egg problem into a sequential, convergent procedure that robustly aligns local audio conditions.

Beyond model design, we developed a scalable pipeline to automatically assemble high-quality, human-centric animation data to address the lack of suitable multi-concept datasets. This pipeline operates by: 1) accurately tracking individual identities to extract their mask information and images, and 2) aligning audio segments to each identity through lip synchronization. We curated a dataset of over two million video-entity pairs, capturing both human-human and human-object interactions across a wide range of object categories. In summary, our contributions are as follows:

{\bf (1)} We propose a novel human animation framework capable of synthesizing multi-person and human-object interactions, conditioned on multiple reference images, text descriptions, and audio inputs. The framework also supports long video generation as well as single full-image conditioning.
{\bf (2)} We highlight the importance of local condition injection for multi-concept, multi-modal video generation and introduce a simple yet effective design that enables the model to handle both global and local conditions by automatically localizing the conditioned layout. Experimental results demonstrate that our proposed design significantly outperforms existing baselines.

\section{Related Works}

\noindent{\bf Video Diffusion Models} have enabled unprecedented quality text-to-video (T2V) or image-to-video (I2V) generation in recent years, thanks to diffusion-based generative models~\citep{jonathan2020ddpm, song2021ddim, karras2022edm, song2020score, liu2022flow}. Early T2V approaches either adapt pretrained text-to-image networks in a training-free manner~\citep{singer2022make, wu2023tune, qi2023fatezero} or fine-tune UNet-based latent diffusion architectures~\citep{guo2024animatediff,zhou2022magicvideo,svd,wang2023modelscope}. To push the frontier, recent works compress spatiotemporal features with 3D causal VAEs~\citep{yu20233DVAE}, and migrate to Diffusion Transformer (DiT) backbones~\citep{vaswani2017attention,peebles2023scalable, videoworldsimulators2024, hong2022cogvideo}. Through progressive low-to-high resolution pretraining and fine-tuning~\citep{polyak2024moviegen, kong2024hunyuanvideo,wang2025wan}, these models yield longer, more coherent, and high-quality videos. Our method is also built upon the pretrained DiT video generation models.

\noindent{\bf Human Animation Models} synthesize videos of people driven by text, reference images, human body poses or audios. Early GAN-based methods~\citep{siarohin2019fomm, zhao2022tps, siarohin2021mraa, jiang2024mobileportrait, wang2021facev2v} are trained on small datasets~\citep{nagrani2017voxceleb, siarohin2019fomm, xie2022vfhq, zhu2022celebv} for self-supervised pose transfer. Diffusion-based approaches~\citep{shao2024human4dit, zhang2024mimicmotion, aa,DBLP:conf/nips/00010ZF0LTCX0L24,wang2025multi} now surpass GAN-based methods by conditioning on 2D skeletons, 3D depth, or mesh sequences. Audio-driven portrait methods~\citep{GeneFace, zhang2023sadtalker, emo, jiang2024loopy, hallo3} have been extended toward full-body motion via two-stage pipelines for improved hand quality~\citep{VLogger, EchomimicV2, EMO2, diffted} and one-stage unified framework designs~\citep{lin2024cyberhost,lin2025omnihuman}. In summary, none of them have explored multi-concept human animation, and our method is the first to enable multi-concept human animation with local audio conditions.

\noindent{\bf Multi-Concept Video Customization Models} have received limited attention. Early single-concept identity-preserving methods include Videobooth~\citep{jiang2024videobooth}, which learns coarse-to-fine embeddings from WebVid~\citep{bain2021frozen} via Grounded-SAM~\citep{kirillov2023segment, liu2023grounding, ren2024grounded}; ID-Animator~\citep{he2024id}, which integrates IP-Adapter~\citep{ye2023ip} into AnimateDiff~\citep{guo2024animatediff}; and ConsisID~\citep{yuan2024identity}, which decouples frequency signals to preserve facial identity. More recently, Video-Alchemist~\citep{chen2025multi}, ConceptMaster~\citep{huang2025conceptmaster}, and Phantom~\citep{liu2025phantom} support multiple reference images and text descriptions via cross- or self-attention injection for general-purpose multi-concept customization. BlobGen-Vid~\citep{feng2025blobgen} further enables local layout control of text-specified concepts using user- or LLM-provided spatiotemporal masks. Ingredients~\citep{fei2025ingredients} proposes to predict the layout of identities in multi-image customization, yet it does not consider the audio condition and does not support multi-person talking generation. All existing methods rely solely on image and text conditions and lack support for multimodal inputs such as audio, which we argue are essential for truly versatile, human-centric video generation.

\section{Method}
\begin{figure}
    \centering
    \includegraphics[width=\textwidth]{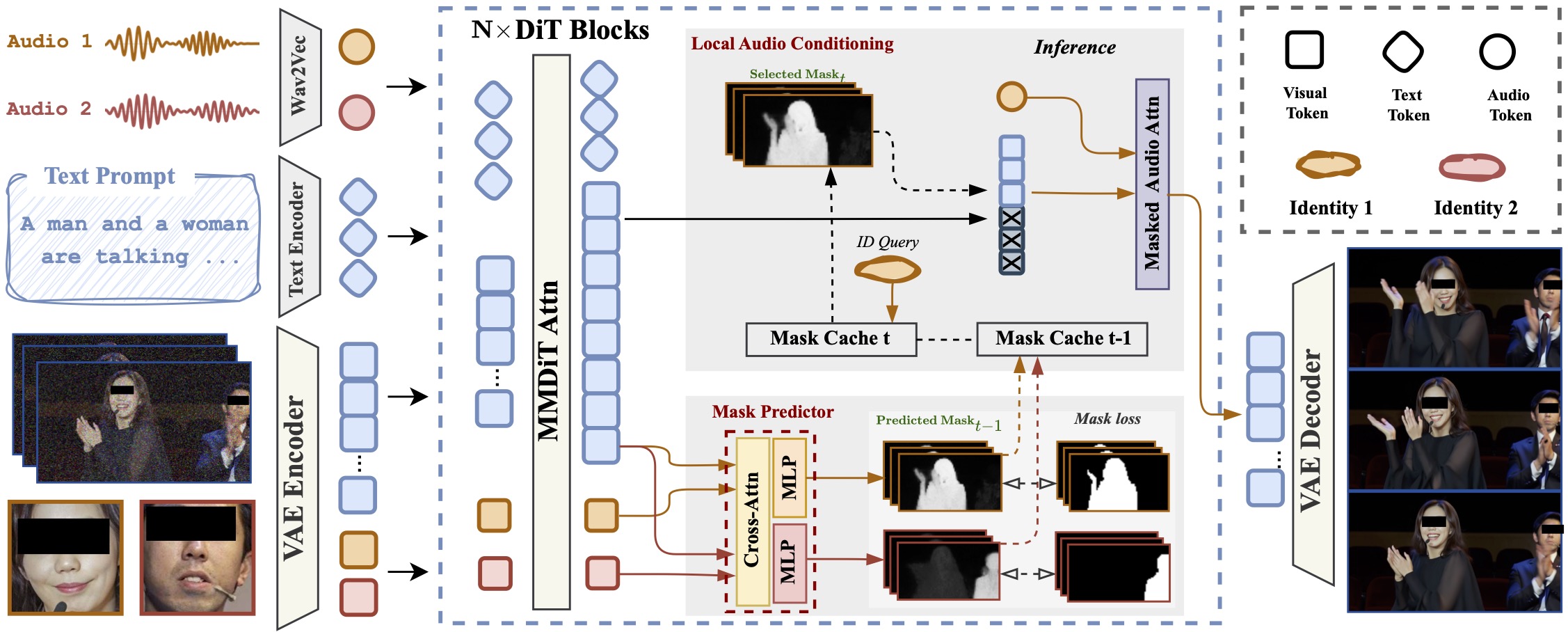}
        \vspace{-2em}
    \caption{Illustration of our framework, which adaptively predicts masks as the spatial guidance of audio condition injection. In training, we train the mask predictor (cross-attn w/ MLP) with mask loss; in inference, we collect mask predictions to cache and leverage masks predicted from the last denoising step ($t-1$) to guide the audio cross-attn in the current denoising step ($t$).}
    \label{fig:frame}
        \vspace{-1em}
\end{figure}

In this section, we present InterActHuman, our multi-concept video generation framework designed to address the challenges of local condition matching for identities in multi-modal conditions. As shown in Fig.~\ref{fig:frame}, the framework begins by customizing reference images of multiple concepts into a video and predicting the mask regions corresponding to each reference appearance in the output video as layout cues with a cross-attention as mask predictor. It operates on the features of the noisy video latent and the reference image latent and supervised by ground-truth masks. Finally, local audio conditions are injected into the designated regions guided by an iterative mask prediction procedure.

\subsection{Preliminaries}
\label{sec:setting}

\noindent{\bf Problem Setting.} Given a caption $T$ that describes a target video, alongside a set of concept reference images $\{ X_i \mid i = 1, \ldots, N \}$ and identity-level audio $\{ Y_i \mid i = 1, \ldots, N \}$ where $N$ denotes the number of distinct concepts, the objective of Multi-Concept Human Animation is to generate high-quality videos that faithfully integrate all image-specified visual concepts with correct lip movements synchronized with each audio signal in accordance with the descriptive caption $T$. Each concept should consistently preserve its visual identity as depicted in the provided images, while accurately expressing the semantic roles and behaviors described in the caption.

\noindent{\bf Preliminaries.} Transformer-based text-to-video latent diffusion models have recently demonstrated remarkable capabilities in generating high-quality video content. Our proposed InterActHuman framework is built upon the MMDiT-based video generation model~\citep{peebles2023scalable,esser2024sd3, seawead2025seaweed} and utilizes a 3D Variational Autoencoder (VAE)~\citep{kingma2013auto}, which compresses input videos into a compact latent space. For training, we adopt the flow matching objective~\citep{lipman2022flow}, which formulates the generative process as a probability flow ordinary differential equation (ODE). This ODE transports clean latent representations $z_0$ to their noisy counterparts $z_t$ along a linear path, defined as $z_t = (1-t)z_0 + t\epsilon$ at timestep $t$, where $\epsilon$ is sampled from a standard Gaussian distribution.  
In our setting, which incorporates multi-modal conditions (image and audio), the output of the diffusion transformer is parameterized as $v_{\Theta}(z_t, t, c_{img}, c_{audio})$. This output is supervised to predict the velocity $(z_1 - z_0)$, resulting in the following training objective:  
\[
\mathcal{L} = \mathbb{E}_{t, z_0, \epsilon} \left|| v_{\Theta}(z_t, t, c_{img}, c_{audio}) - (z_1 - z_0) \right||_2^2.
\]  
This formulation ensures the model effectively learns to generate videos conditioned on multi-modal inputs while maintaining high fidelity and temporal consistency.

\noindent{\bf Multi-Concept Reference Image Injection.}
To handle multiple appearance images, we inject appearance cues via self-attention in the original DiT layers. Each reference image $X_i$ is encoded into latent tensors ${\mathbf{x}_i}$ using the same VAE as for noisy video frames. The reference latents are then flattened into token sequences and processed by the DiT along with the noisy latents $\mathbf{v}$. The reference latents will reuse the parameters of DiT to extract features and interact with noisy latents at the self-attention layer in every DiT block, allowing appearance cues to propagate to the denoising pathway. Notably, no extra networks or additional parameters are required, thus preserving the model's efficiency. Please refer to Appendix \ref{sec:omnihuman_details} for more details.

\subsection{Layout Prediction and Local Condition Injection}
While reference-image injection is well studied, leveraging spatiotemporal layouts for local conditions, such as precise audio alignment, remains challenging. Mask prediction and audio injection are coupled: if a denoising step is incomplete, the masks are unavailable; if it is complete, it is too late to inject local conditions. We resolve this with the multi-step inference of diffusion models~\citep{song2021ddim}: masks predicted at step $k$ guide multi-modal condition injection at step $k{+}1$. As the network learns spatiotemporal layouts for each identity, precise masks emerge without user annotations.

\textbf{Mask Predictor.} In each DiT block, we predict a spatiotemporal mask quantifying how strongly each reference image should influence each video frame. A lightweight mask-predictor head is attached to each of the $L$ transformer layers, consisting of: (1) a shared linear projection that maps hidden video features $\mathbf{h}^{v}$ and hidden reference features $\mathbf{h}^{r}_i$ to query, key, and value tensors; (2) LayerNorm; (3) 3D RoPE positional encoding; (4) a cross-attention module; and (5) a two-layer MLP. After normalization and RoPE, video tokens attend to a single reference via $a_i^{(l)} = \operatorname{softmax}\left(\frac{\mathbf{Q}^{v}{\mathbf{K}^{r}_{i}}^{\top}}{\sqrt{d}}\right)\mathbf{V}^{r}_{i}$, 
where $d$ is the head dimension and $l$ indexes the current layer. The attended feature $\mathbf{a}^{(l)}_i$ is transformed by the MLP and passed through a sigmoid to yield a layer-specific mask $\mathbf{m}^{(l)}_i \in [0,1]^{T}$ for reference $X_i$. We then average the predictions from the last few DiT blocks to obtain the final mask. Notably, the mask predictor is trained to recover the complete human region, regardless of whether the reference image shows only the upper body, face, or full body. This simplifies mask prediction and stabilizes the conditioning module, yielding robust behavior across diverse scenarios. Because the predictors reuse in-block features and share parameters across references, they add minimal overhead while enabling explicit layout control for each identity.

\noindent \textbf{Mask Prediction by Caching During Inference.} During inference, aggregating mask signals across layers is difficult in early denoising steps, when reference concept latents are still ambiguous. We adopt an iterative strategy: at each step, the mask predicted at the previous step is cached and used as a layout prior for the current step. This progressively sharpens spatiotemporal localization, allowing masks to adapt to the evolving content and remain consistent per concept over time, thereby improving overall conditioning quality.

\noindent \textbf{Local Audio Conditioning.} To prepare for multi-concept, human-centric video generation, we first pretrain on single-identity audio-conditioned animation by adding cross-attention–based audio conditioning (without masks) and a mixed-conditions training strategy, following~\citep{lin2025omnihuman}. Concretely, a new cross-attention layer injects wav2vec~\citep{baevski2020wav2vec2} audio features into each DiT block after the MMDiT layer. In the multi-concept setting, this pretrained audio cross-attention already provides strong per-identity audio control. We therefore implement \emph{local} audio conditioning primarily at inference: instead of updating all noisy video tokens~\citep{lin2025omnihuman,lin2024cyberhost}, we inject wav2vec features only into tokens assigned to a given identity, using the previous-step mask as guidance. To ensure smooth transitions in latent features and in the final video, we blend meaningful and muted audio features per token using the mask confidence, with soft weighting near mask boundaries. This local conditioning also enables multi-person dialogues. Given per-speaker audio tracks as input, the model generates realistic interactions in which speakers take turns, supporting a wide range of applications.

\noindent \textbf{Training Loss and Strategies.} The overall objective combines a flow-matching diffusion loss (Sec.~\ref{sec:setting}) with a focal loss~\citep{lin2017focal} for mask classification, which stabilizes training compared to binary cross-entropy, likely due to foreground–background imbalance and occasional low-quality masks. A frame-alignment flag excludes frames with invalid or low-quality masks from loss computation, strengthening temporal supervision. To mitigate the common “copy–paste” behavior of diffusion models—replicating subjects from references with little variation in pose or viewpoint—we randomly mask reference images to reveal only the head, full body, or clothing, promoting appearance diversity. We use a face-to–full-body appearance conditioning ratio of $0.7:0.3$ and weight the diffusion and focal losses equally ($1:1$).

\noindent \textbf{Inference Strategies.} Given $N$ reference images and a text prompt, our method uses Qwen2.5-VL~\citep{bai2025qwen2} as a rephraser to extract detailed descriptions from each reference image and integrate them with the original prompt. We apply a shared classifier-free guidance (CFG) across audio and text. For local control, audio conditioning is applied only within the predicted mask regions for each reference. Because early denoising steps yield unreliable masks that may suppress other references, we disable mask-based injection for the first 10 steps, then enable it using the previous-step mask. We inject masked audio only on the conditional (positive) branch during CFG sampling, with a CFG scale of 6.5 and 50 denoising steps.

\begin{figure}[t]

    \centering
    \includegraphics[width=\textwidth]{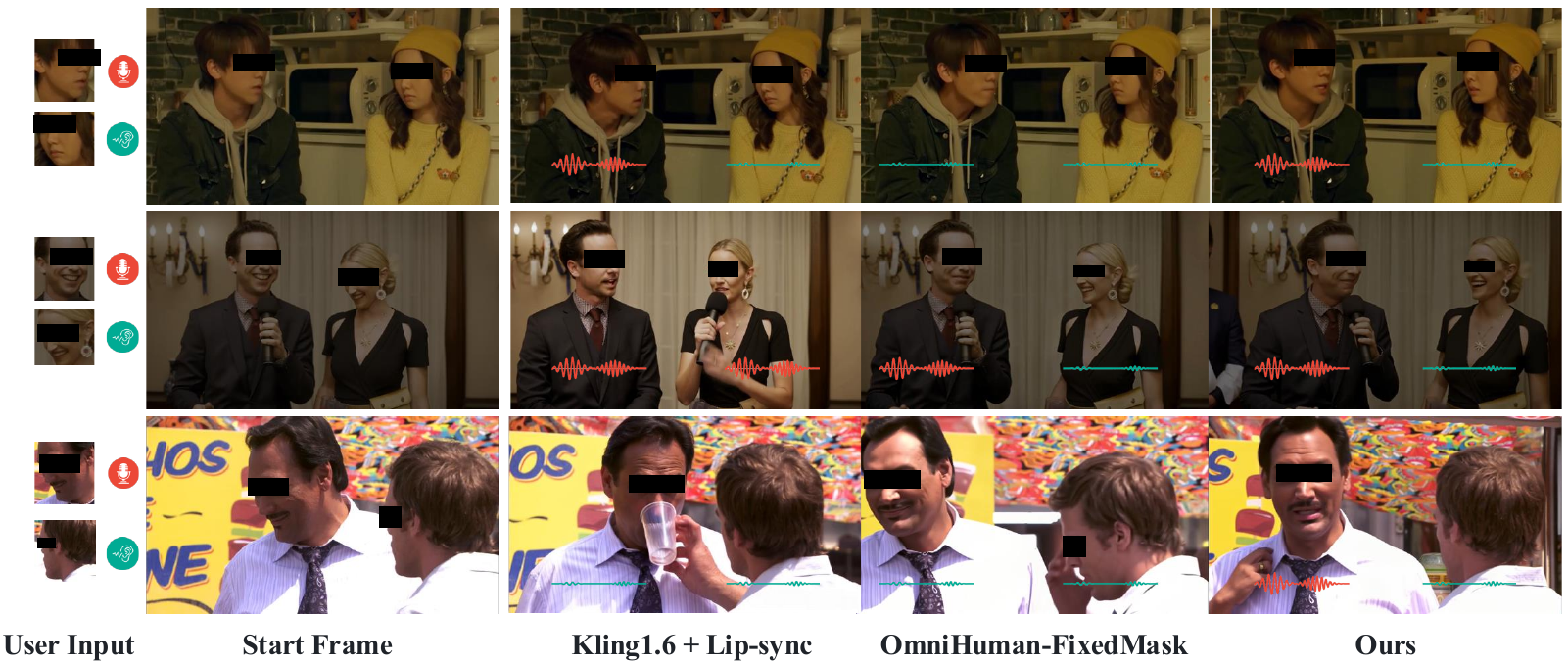}

        \vspace{-1em}
    \caption{Qualitative comparison with previous methods on multi-concept audio injection. }
            \label{fig:audio}
            \vspace{-1em}
\end{figure}

\subsection{Multi-Concept Human-Centric Video Data Curation}

We curate a large-scale, human-centric video dataset from a large public video dataset~\citep{li2024openhumanvid} and self-collected videos, filtering out videos that are too short ($<$4 seconds) or contain too many salient humans ($>$3) as detected by a human pose detector~\citep{dwpose}. Unlike previous multi-concept video generation methods~\citep{chen2025multi,huang2025conceptmaster} that rely on conventional pipelines like Grounding-DINO~\citep{liu2023grounding} and SAM~\citep{kirillov2023segment} for obtaining a single reference image per identity, our multi-stage pipeline leverages advanced vision-language models. First, each raw video is densely captioned using Qwen2-VL~\citep{wang2024qwen2} (distilled from Gemini-2.0-Pro), generating fine-grained descriptions of the environment, subjects’ appearance, actions, expressions, interacting salient objects, and inter-subject interactions. Next, these descriptions are parsed by a zero-shot Gemini-2.0-Flash API to extract structured appearance phrases. For spatial supervision, we use Grounding-SAM2~\citep{ren2024grounded} with the query \texttt{person} to produce accurate, temporally consistent masks, which are used both for extracting foreground reference images (with a white background) and as ground-truth for our mask predictor. Our corpus comprises over $2.6$M triplets of videos, per-frame masks, and captions, forming the foundation of InterActHuman.

\section{Experiments}

\noindent \textbf{Baselines.} Since ConceptMaster~\citep{huang2025conceptmaster} and Video-Alchemist~\citep{chen2025multi} are not open-sourced, we do not have access to their models nor test sets. Therefore, we primarily compare our method with recent multi-concept video customization models through their publicly available APIs or public models, evaluating performance from the perspectives of visual appearance and adherence to text prompts, including Vidu2.0~\citep{bao2024vidu}, Pika2.1~\citep{pikaart2024}, Kling 1.6~\citep{kuaishou2024} and Phantom~\citep{liu2025phantom}, and audio-driven human video generation methods, including DiffTED~\citep{diffted}, DiffGest~\citep{zhu2023taming} + Mimiction~\citep{zhang2024mimicmotion}, CyberHost \citep{lin2024cyberhost}, OmniHuman~\citep{lin2025omnihuman} and Kling 1.6~\citep{kuaishou2024} with lip synchronization. 

\begin{table}
\caption{Quantitative comparisons with audio-conditioned full-body animation baselines.}
\centering
  \resizebox{0.95\textwidth}{!}{
\begin{tabular}{lccc|ccccc}
\toprule
\multirow{2}{*}{Methods} & \multicolumn{3}{c}{Single-Person Test Set} & \multicolumn{4}{c}{Multi-Person Test Set} \\
\cmidrule(lr){2-4} \cmidrule(lr){5-8}
 & Sync-C$\uparrow$ & HKV$\uparrow$ & HKC$\uparrow$ & Sync-D$\downarrow$ & IQA$\uparrow$ & AES$\uparrow$ & FVD$\downarrow$ \\
\midrule
DiffTED & 0.926 & - & 0.769 & - & - & -  & - \\ 
DiffGest.+Mimic. & 0.496 & 23.409 & 0.833 & - & - & -  & - \\ 
CyberHost & 6.627 & 24.733 & 0.884 & 8.974 & 4.011 & 2.856 &  54.797 \\
Kling1.6 + Lip-sync. & 4.449 & 46.490 & 0.826 & 8.401 & 4.716 & 3.444 & 33.555 \\
MultiTalk & - & -& - & 7.671 & 4.561 & 3.248 & 35.472 \\ 
OmniHuman w/o mask & \textbf{7.443} & 47.561 & \textbf{0.898}& 9.482 & \textbf{4.768} & 3.466 & 33.895 \\  
OmniHuman w/ fixed mask & - & -& - & 7.068 & 4.690 & 3.369  & 40.239 \\  
\midrule
Ours & 7.272 & \textbf{59.635} & 0.885 & \textbf{6.670} & 4.757 & \textbf{3.467} & \textbf{22.881} \\
\bottomrule
\end{tabular}}
\label{tab:audio}
\vspace{-1em}
\end{table}

\begin{table}[t]
    \caption{User preference evaluation. $^\star$ means publicly available version with Wan2.1-1.3B.}
    \label{tab:user_eval}
    \centering
    \resizebox{0.99\textwidth}{!}{
    \begin{tabular}{lccc|ccccc}
        \toprule
         & \multicolumn{3}{c}{Audio-Driven} & \multicolumn{5}{c}{Multi-Concept Customization} \\
         \cmidrule(lr){2-4} \cmidrule(lr){5-9}
        Metric & Kling & OmniHuman & Ours 
        & Pika & Phantom$^\star$ & Kling & Vidu & Ours \\
        \midrule
        Avg. Score $\uparrow$ & 1.70 & 1.82 & \textbf{2.48} & 2.22 & 2.46 & 2.90 & 3.40 & \textbf{4.01} \\
        Top-1 (\%) $\uparrow$ & 14.5\% & 25.6\% & \textbf{59.9\%} & 4.9\% & 9.9\%& 13.6\% & 22.2\%  & \textbf{49.4\%} \\
        \bottomrule
    \end{tabular}
    } 
        \vspace{-1em}
\end{table}

\begin{table}[t]
	\centering
    \caption{Quantitative comparison of subject consistency, prompt following and visual quality. $^\star$ means publicly available version with Wan2.1-1.3B.}
	\resizebox{\textwidth}{!}{
		\small
		\begin{tabular}{lccccccccc}
			\toprule
			\multirow{2}{*}{Methods} & \multicolumn{5}{c}{Decoupled Concept Fidelity} & \multicolumn{1}{c}{Prompt} & \multicolumn{3}{c}{Video Quality}\\
			\cmidrule(lr){2-6} \cmidrule(lr){7-7} \cmidrule(lr){8-9}
			& CLIP-I$\uparrow$ & DINO-I$\uparrow$ & Face-Arc$\uparrow$ & Face-Cur$\uparrow$ & Face-Glink$\uparrow$ & ViCLIP-T$\uparrow$ & AES$\uparrow$ & IQA$\uparrow$  \\ 
			\midrule
			Vidu2.0 & 0.696 & 0.458 & 0.568 & 0.562 & 0.597 & 18.61 & 3.350 & 4.689 \\ 
			Pika2.1 & 0.688 & 0.459 & 0.579 & 0.566 & 0.607 & \textbf{19.39}  & 3.534 & 4.791\\ 
			Kling1.6 & 0.659 & 0.420 & 0.552 & 0.547 & 0.582 & 18.38 & 3.487 & 4.787 \\ 
			Phantom$^\star$ & 0.703 & 0.476 & 0.589 & 0.573 & 0.615 & 17.73  & 3.404 & 4.812 \\ 
            
            \midrule
            Ours & \textbf{0.744} & \textbf{0.533} & \textbf{0.598} & \textbf{0.600} & \textbf{0.644} & 18.87 & \textbf{3.565} & \textbf{4.903}  \\
			\bottomrule
		\end{tabular}
	}
        
	\label{tab:img}
    \vspace{-1em}
\end{table}

\begin{figure}[t]
    \centering
    \includegraphics[width=\textwidth]{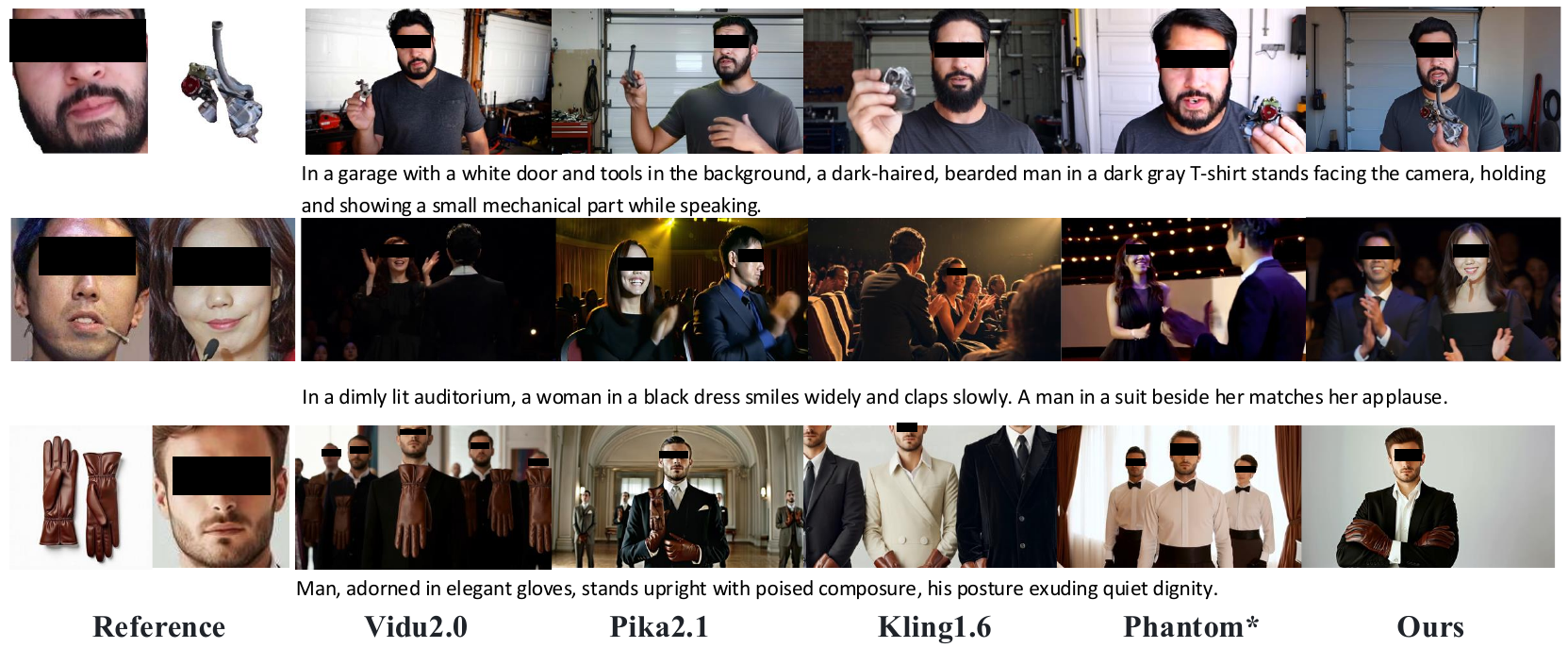}
        \vspace{-1em}
    \caption{Qualitative comparison with previous methods on subject consistency and text following.}
        \label{fig:id}
\vspace{-1em}
\end{figure}

\noindent \textbf{Evaluation Metrics.} As {\em audio} and {\em appearance} are key characteristics of humans, we follow current state-of-the-art methods~\citep{lin2025omnihuman,huang2025conceptmaster} to evaluate audio-driven, multi-concept human-centric video generation. To comprehensively evaluate methods for this task, we consider five dimensions:
{\em 1) Concept fidelity}: We leverage image feature extractors like CLIP-I~\citep{radford2021learning} and DINO-I~\citep{oquab2023dinov2} on subjects in the generated videos to assess whether they are aligned with the provided reference images. To get cropped subject images in output videos, we randomly sample 5 frames in each video and use Florence-2~\citep{xiao2024florence} to detect the subject bounding boxes. We then crop the detected regions and compute the CLIP-I and DINO-I scores. We also employ Face-Arc~\citep{deng2019arcface}, Face-Cur~\citep{huang2020curricularface} and Face-Glink~\citep{deng2019arcface} to evaluate the fidelity of facial features if the concept is a human.
{\em 2) Prompt following}: We utilize video-level CLIP~\citep{wang2022internvideo} to measure the similarity between the input text prompts and the resulting video content.
{\em 3) Visual quality}: We utilize q-align~\citep{wu2023q}, a vision-language model, for no-reference image quality assessment (IQA) and aesthetic score estimation (AES).
{\em 4) Audio synchronization and human pose diversity}: For lip synchronization, we leverage the widely used Sync-C and Sync-D~\citep{syncnet} to compute audio-visual confidence. We also incorporate hand keypoint confidence (HKC) and hand keypoint variance (HKV)~\citep{lin2024cyberhost} to quantify hand-pose accuracy and motion richness, respectively. For simplicity, we assume each test video contains exactly two participants: one speaking (with meaningful audio as input) and one listening (with muted audio as input).
{\em 5) Distribution distance}: We employ FVD~\citep{unterthiner2019fvd} to measure the distance between generated and ground-truth videos. 

\noindent \textbf{Test Sets.} We use three test sets in our experiments: 1) single-person audio conditioned human animation test set following OmniHuman~\citep{lin2025omnihuman} (see Tab.~\ref{tab:audio}); 2) our collected two-person audio conditioned human animation test set where only one person is talking (see Tab.~\ref{tab:audio} and Tab.~\ref{tab:ablation}); and 3) multi-concept video customization test set following Phantom~\citep{liu2025phantom}, where we select 100 human-related pairs of reference images and text prompts in our experiments (see Tab.~\ref{tab:img}).

\subsection{Comparisons with State-of-the-Arts}
\noindent \textbf{Multi-Concept Audio-Driven Human Video Generation.}
In Tab.~\ref{tab:audio}, our approach achieves state-of-the-art or comparable performance in lip synchronization accuracy and motion diversity, particularly excelling in complex multi-person interactions where baseline methods~\citep{lin2024cyberhost,kuaishou2024} struggle with accurate audio signal assignments. For single-person scenarios, our method performs on par with specialized models like OmniHuman~\citep{lin2025omnihuman}. In multi-person settings, existing methods including OmniHuman and its extensions as well as leading commercial video generation models with post-processed lip-sync, fail to deliver satisfactory results. Poor lip-sync accuracy highlights their inability to generate accurate lip movements for the correct person. Although OmniHuman with oracle masks (manually setting audio-conditioned regions) improves lip-sync accuracy, it still significantly degrades overall video quality, as reflected in the FVD metric. This clearly demonstrates the limitations of existing methods that rely on single-identity assumptions, hindering their applicability to broader scenarios.
In contrast, our method achieves strong performance in both lip-sync accuracy and overall video synthesis quality, effectively addressing the challenges of audio-conditioned multi-person video generation while remaining compatible with existing settings.
Fig.~\ref{fig:audio} presents qualitative results for multi-person audio-driven video generation. While Kling1.6 w/ lip-sync shows many audio assignment errors and OmniHuman w/ fixed mask shows inflexible audio control with many missing cases, our method consistently assigns audio signals to the correct identity and demonstrates better motion dynamics and more precise audio-driven animation. It is worth noting that all previous methods rely on a reference frame containing complete information to generate talking-person videos, while our method only needs reference appearance of human's head or full-body images and audios.

\noindent \textbf{User Study.} We conducted a user study to evaluate our method on two tasks: (1) lip synchronization in multi-person talking videos and (2) subject consistency in multi-concept customizations. The lip sync test used 19 videos from three methods, while subject consistency was assessed on 9 videos from five methods. We use the same model (labeled `ours') for two evaluation tasks. Ten experienced users ranked each method with scores (higher values indicate superior performance). Tab.~\ref{tab:user_eval} reports the average scores and top-1 selection percentages. Our method achieved the highest scores and top-1 rates in both tasks, validating the effectiveness of our method.

\begin{table}[t]
\caption{Ablation study on audio-driven multi-person animation methods.}
    \centering
    \resizebox{0.6\textwidth}{!}{
    \begin{tabular}{lcccc}
        \toprule
        \textbf{Variants} & Sync-D$\downarrow$ & IQA$\uparrow$ & AES$\uparrow$ & FVD$\downarrow$ \\
        \midrule
        Global audio condition & 9.482 & \textbf{4.768} & 3.466 & 33.895 \\
        ID Embedding & 8.627 & 4.658 & 3.338 & 35.665 \\
        Fixed Mask & 7.068 & 4.690 & 3.369 & 40.239 \\
        \midrule
        Predicted Mask (Ours) & \textbf{6.670} & 4.757 & \textbf{3.467} & \textbf{22.881} \\
        \bottomrule
    \end{tabular}}
    \label{tab:ablation}
    \vspace{-1em}
\end{table}

\begin{table}[t]
\centering
\caption{Runtime and parameters versus number of reference images.}
\label{tab:runtime}
\begin{tabular}{lccc|c}
\toprule
Component & 1 ref & 2 refs & 3 refs & \#Params \\
\midrule
DiT Model & 6.5s & 7.0s & 7.7s & 7B \\
Mask Predictor & 0.4s & 0.8s & 1.2s & 56M \\
Full Model & 6.9s & 7.8s & 8.9s & 7B \\
\bottomrule
\end{tabular}
\vspace{-1em}
\end{table}

\begin{figure}[t]
    \centering
    \includegraphics[width=0.8\textwidth]{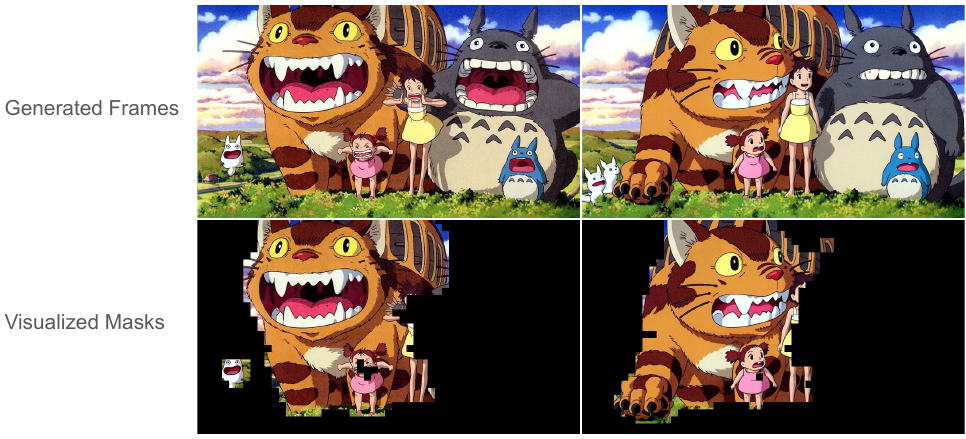}
    \vspace{-1em}
    \caption{Qualitative results on the failure case of mask predictions.}
        \label{fig:failure-mask}
        \vspace{-1em}
\end{figure}

\noindent \textbf{Multi-Concept Video Customization.}
Although our major contribution is not on multi-concept video customization, we show that our model is also capable of preserving multi-concept visual appearances.
In Tab.~\ref{tab:img}, our method outperforms existing approaches~\citep{bao2024vidu,pikaart2024,kuaishou2024} in preserving identity details and facial features, addressing the common degradation issue in multi-subject generation. Notably, we achieve this without sacrificing audio injection performance, while these methods are incapable of generating videos from audio conditions. It suggests our joint optimization framework on video generation and mask prediction successfully balances video generation. Our method ranks second in prompt following, likely due to our audio-driven human-centric training data focusing on talking and singing, which limits prompt diversity compared to text-to-video tasks. Despite this, qualitative results (Fig.~\ref{fig:id}) show that our approach maintains natural subject consistency and visual quality in both real-world and anime domains, outperforming previous methods that suffer from unnatural compositions and degraded visuals.

\subsection{Ablation Study}
\begin{figure}[t]
    \centering
    \includegraphics[width=\textwidth]{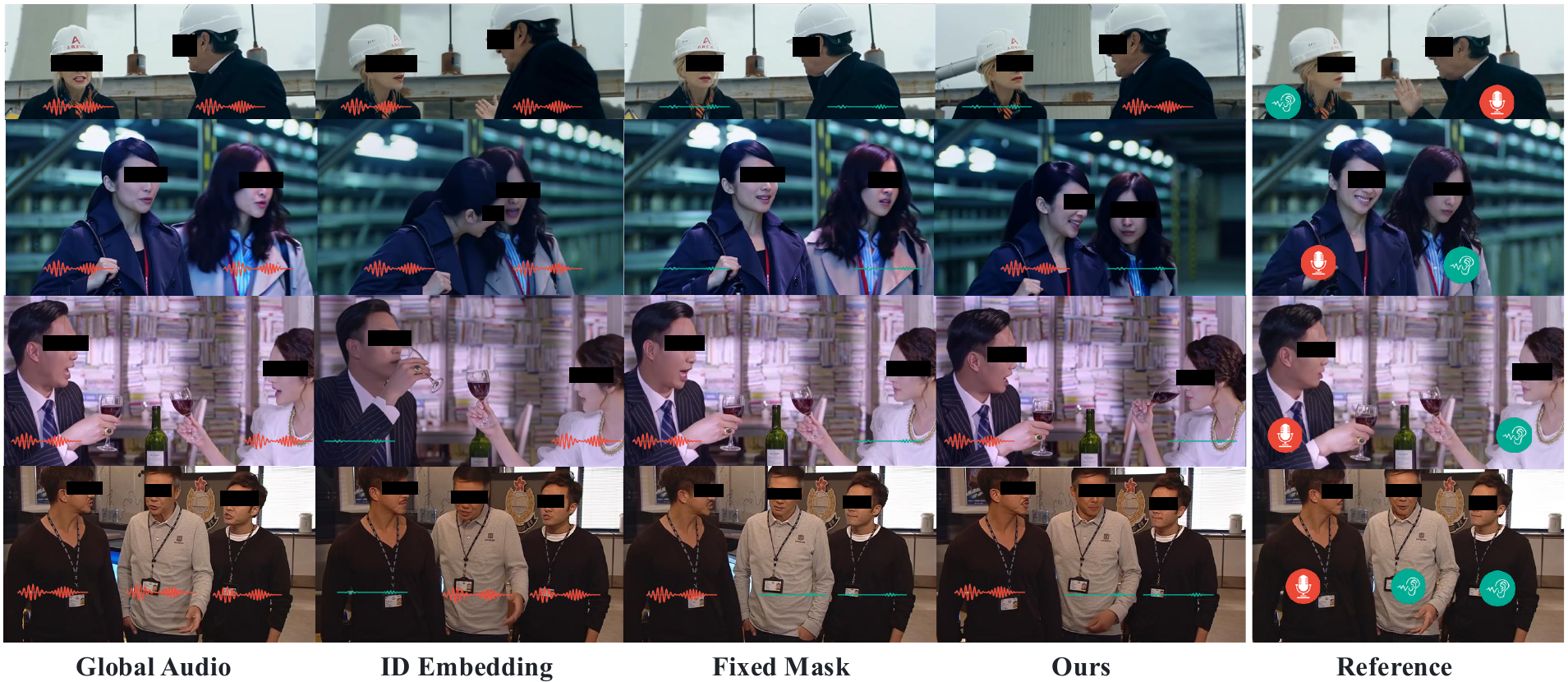}
    \vspace{-1em}
    \caption{Qualitative ablation on audio injection strategies.}
        \label{fig:ablation}
        \vspace{-1em}
\end{figure}

We validate our local audio injection designs via ablation on three variants: \emph{1) Global audio}~\citep{lin2025omnihuman} applies audio across the entire feature map;
\emph{2) ID Embedding} injects audio features with a learnable ID embedding without mask prediction;
\emph{3) Fixed Mask} uses predefined static spatial masks.
As shown in Tab.~\ref{tab:ablation}, our framework with predicted dynamic masks achieves the best Sync-D and FVD scores. The fixed mask yields decent Sync-D but suffers motion artifacts (worst FVD), and global audio, while scoring best in IQA, offers poor audio-visual alignment. Qualitative results in Fig.~\ref{fig:ablation} further reveal that global audio drives all identities uniformly, ID embedding often mismatches audio with identities, and fixed masks lose alignment when characters move. These findings underscore that current methods overlook the need for precise local conditions, highlighting the strength of our dynamic, adaptive mask prediction strategy for multi-person talking video generation.

\noindent {\bf Computational Cost and Runtime}
\label{app:runtime}
We measure inference time with the setting of 720p, 109 frames (28 VAE latents) on single A100, and record single forward pass time. The mask predictor adds \(\sim\)56M parameters versus a 7B DiT and incurs a small overhead per reference image (\(\approx\)0.013s per DiT block). As more references increase VAE latents from \(28\) to \(28{+}n\), DiT self-attention scales as \((1{+}n/28)^2\).

\noindent {\bf Failure case analysis of mask prediction}
\label{app:failure-mask}
In Fig.~\ref{fig:failure-mask} we show the qualitative results on the failure case of mask predictions, where the yellow cat is the current identity image (i.e., ID query of mask predictor). We can see that although the mask prediction achieves good performance in IoU measurement from Tables~\ref{tab:mask_low} and  \ref{tab:mask_high}, it still makes mistakes when there are highly overlapping regions. For example, two girls are standing in front of the yellow cat, and the mask predictor manages to exclude the upper body of the right girl in yellow because of small area of overlapping regions. Yet, the mask predictor fails for the girl in pink in both frames. In the first frame, the mask predictor manages to exclude the face region of the little girl in pink, but it mistakenly include another white animal. Besides, due to the high VAE down-sampling ratio of the widely used DiT models, the mask is predicted in a very low resolution, leading to inaccurate mask boundaries.

\noindent {\bf Additional analyses in Appendix.}
We include (i) a capability matrix against concurrent systems (Tables~\ref{tab:capability}), (ii) mask IoU versus denoising steps/layers under low/high motion plus motion statistics (Tables~\ref{tab:mask_low}, \ref{tab:mask_high}, \ref{tab:motion_stats}), (iii) a mask cache ablation (Table~\ref{tab:mask_cache}), (iv) scaling to $>$3 people (Table~\ref{tab:more_people}).

\section{Conclusion}
In this paper, we introduced InterActHuman, a novel end-to-end video diffusion framework that supports spatially aligned, multi-modal conditioning for multi-concept human animation. By integrating an efficient mask-prediction module into a pretrained DiT backbone, our method automatically infers per-identity spatio-temporal layouts from reference images and uses these masks to guide local audio conditioning. This explicit layout binding enables each reference concept to retain its unique appearance and voice, even in complex scenes with multiple interacting entities. To facilitate training procedure of our model, we presented a high-throughput pipeline for harvesting and annotating over 2.6 million identity-aware video snippets, complete with per-frame masks for humans and objects. Extensive experiments on both single- and multi-person benchmarks demonstrate that InterActHuman achieves state-of-the-art performance in lip synchronization, motion diversity and subject appearance fidelity, while maintaining competitive video quality. We hope it could serve as a solid baseline for the multi-concept human animation and audio-driven multi-person video generation community, where further improvements could be built upon it.

\noindent {\bf Limitations.} 
Since the goal of this work is for human-centric video generation, the available data domain is inherently narrower than that used for general text-to-video pretraining, limiting the ability to follow the diverse text prompts. While our framework is designed to accommodate any number of concept images, the training dataset predominantly consists of videos featuring two to three individuals. This distribution may constrain the generalization of any number of inputs, suggesting further improvements could be achieved by enlarging the diversity and scale of training data.

\noindent {\bf Acknowledgment}. We thank the infrastructure team at ByteDance Intelligent Creation for the data curation and annotation support.

\noindent {\bf Ethics Statement.} Our model could be used to generate misinformation by using celebrity images and voices. We will strictly restrict access and add watermarks to prevent misuse.

\noindent {\bf Reproducibility Statement.} We have provided a code re-implemented on the publicly available video diffusion pretraining model Wan2.1~\citep{wang2025wan} to show the details of our method. We also provided the dataset processing code, e.g., caption analysis and Grounding-SAM2~\citep{ren2024grounded}. In the Algorithm~\ref{alg:interacthuman}, we provided detailed pseudo-code to show the inference pipeline of our method. Based on such information, we believe the reproducibility of method is solid enough.
\clearpage
\bibliography{iclr2026_conference}
\bibliographystyle{iclr2026_conference}
\appendix

\section{Additional Experimental Results}

\subsection{LLM Usage Statement.} We use LLMs (e.g., Gemini-2.5 and GPT-5) to polish our paragraphs.

\subsection{Comparison with Single-Person Talking-Head Methods}
\label{app:talkinghead}
Although our method targets multi-person \emph{full-body} talking video generation, we report single-person \emph{talking-head} results for completeness.

\begin{table}[h]
\centering
\caption{CelebV-HQ: higher is better for IQA/ASE/Sync-C; lower is better for FID.}
\label{tab:celebv_table}
\begin{tabular}{lcccc}
\toprule
Method & IQA$\uparrow$ & ASE$\uparrow$ & Sync-C$\uparrow$ & FID$\downarrow$ \\
\midrule
SadTalker~\citep{zhang2023sadtalker} & 2.953 & 1.812 & 3.843 & 36.648 \\
Hallo~\citep{xu2024hallo} & 3.505 & 2.262 & 4.130 & 35.961 \\
VExpress~\citep{wang2024vexpress} & 2.946 & 1.901 & 3.547 & 65.098 \\
EchoMimic~\citep{chen2024echomimic} & 3.307 & 2.128 & 3.136 & 35.373 \\
Hallo-3~\citep{hallo3} & 3.451 & 2.257 & 3.933 & 38.481 \\
Loopy~\citep{jiang2024loopy} & 3.780 & 2.492 & 4.849 & 33.204 \\
OmniHuman~\citep{lin2025omnihuman} & \textbf{3.875} & \textbf{2.656} & \textbf{5.199} & 31.435 \\ \midrule
\textbf{Ours} & 3.834 & 2.553 & 5.047 & \textbf{30.159} \\
\bottomrule
\end{tabular}
\end{table}

As shown in Table~\ref{tab:celebv_table}, OmniHuman achieves the best single-person talking-head scores on IQA/ASE/Sync-C, while {\em our method yields the best FID}. Since our approach targets multi-person, full-body generation rather than single-person heads, these results indicate competitive quality under a setting that is not our primary target. On RAVDESS (Table~\ref{tab:ravdess_table}), our method attains the best IQA and ASE, while OmniHuman leads on Sync-C and FID. This mirrors the CelebV-HQ trend and supports our claim that the proposed framework remains competitive on single-person benchmarks while being designed for multi-person, full-body scenarios.

\begin{table}[h]
\centering
\caption{RAVDESS: higher is better for IQA/ASE/Sync-C; lower is better for FID.}
\label{tab:ravdess_table}
\begin{tabular}{lcccc}
\toprule
Method & IQA$\uparrow$ & ASE$\uparrow$ & Sync-C$\uparrow$ & FID$\downarrow$ \\
\midrule
SadTalker~\citep{zhang2023sadtalker} & 3.840 & 2.277 & 4.304 & 32.343 \\
Hallo~\citep{xu2024hallo} & 4.393 & 2.688 & 4.062 & 19.826 \\
VExpress~\citep{wang2024vexpress} & 3.690 & 2.331 & 5.001 & 26.736 \\
EchoMimic~\citep{chen2024echomimic} & 4.504 & 2.742 & 3.292 & 21.058 \\
Hallo-3~\citep{hallo3} & 4.006 & 2.462 & 4.448 & 28.840 \\
Loopy~\citep{jiang2024loopy} & 4.506 & 2.658 & 4.814 & 17.017 \\
OmniHuman~\citep{lin2025omnihuman} & 4.564 & 2.815 & \textbf{5.255} & \textbf{16.970} \\ \midrule
\textbf{Ours} & \textbf{4.602} & \textbf{2.915} & 5.132 & 17.187 \\
\bottomrule
\end{tabular}
\end{table}

\subsection{Comparison with concurrent works}
\label{app:capability}
We show comparison with recent multi-person talking systems which are conditioned on a \emph{reference frame} only and released concurrently with our method. Our framework supports both concept-only and first-frame modes.

\begin{table}[h]
\centering
\caption{Qualitative capability comparison. $\checkmark$: supported; \textbf{x}: not supported.}
\label{tab:capability}
  \resizebox{\textwidth}{!}{
\begin{tabular}{lcccc}
\toprule
Methods & Audio-driven full-body & First-frame I2V & Multi-ref images & Multi-person talking \\
\midrule
Video-Alchemist~\citep{chen2025multi} & \textbf{x} & \textbf{x} & $\checkmark$ & \textbf{x} \\
ConceptMaster~\citep{huang2025conceptmaster} & \textbf{x} & \textbf{x} & $\checkmark$ & \textbf{x} \\
Phantom~\citep{liu2025phantom} & \textbf{x} & \textbf{x} & $\checkmark$ & \textbf{x} \\
OmniHuman~\citep{lin2025omnihuman} & $\checkmark$ & $\checkmark$ & \textbf{x} & \textbf{x} \\
HunyuanVideo-Avatar~\citep{chen2025hunyuanvideo} & $\checkmark$ & $\checkmark$ & \textbf{x} & $\checkmark$ \\
MultiTalk~\citep{kong2025let} & $\checkmark$ & $\checkmark$ & \textbf{x} & $\checkmark$ \\
\textbf{Ours} & $\checkmark$ & $\checkmark$ & $\checkmark$ & $\checkmark$ \\
\bottomrule
\end{tabular}}
\end{table}

\subsection{Mask Refinement Across Denoising Steps and Layers}
\label{app:mask_iou}
Across both regimes (Tables~\ref{tab:mask_low}, \ref{tab:mask_high}), IoU steadily improves with denoising steps, validating the \emph{iterative refinement} strategy. Deeper layers (e.g., layer 36) outperform shallow layers early on, while the \emph{combined} mask yields the strongest mid/late-step IoUs. Notably, high-motion sequences retain strong IoUs and achieve competitive lip-sync (Sync-D \textbf{6.921}), indicating that masks are \emph{not static} and remain reliable even under larger motion (see the motion statistics in Table~\ref{tab:motion_stats}).

\begin{table}[h]
\centering
\caption{Low motion strength: mask IoU across steps/layers; Sync-D for the setting.}
\label{tab:mask_low}
\begin{tabular}{lccccc|c}
\toprule
 & step 1 & step 10 & step 20 & step 30 & step 50 & Sync-D$\downarrow$ \\
\midrule
Mask layer 4  & 0.306 & 0.436 & 0.509 & 0.640 & 0.858 & -- \\
Mask layer 20 & 0.386 & 0.629 & 0.733 & 0.890 & \textbf{0.957} & -- \\
Mask layer 36 & \textbf{0.538} & \textbf{0.742} & 0.850 & 0.916 & 0.923 & -- \\
Combine mask  & 0.376 & 0.738 & \textbf{0.881} & \textbf{0.949} & 0.956 & \textbf{7.292} \\
\bottomrule
\end{tabular}

\end{table}

\begin{table}[h]
\centering
\caption{High motion strength: mask IoU across steps/layers; Sync-D for the setting.}
\label{tab:mask_high}
\begin{tabular}{lccccc|c}
\toprule
 & step 1 & step 10 & step 20 & step 30 & step 50 & Sync-D$\downarrow$ \\
\midrule
Mask layer 4  & 0.113 & 0.426 & 0.529 & 0.730 & 0.779 & -- \\
Mask layer 20 & 0.527 & 0.816 & 0.911 & 0.934 & \textbf{0.945} & -- \\
Mask layer 36 & 0.694 & 0.902 & 0.923 & 0.931 & 0.915 & -- \\
Combine mask  & \textbf{0.741} & \textbf{0.916} & \textbf{0.932} & \textbf{0.936} & 0.937 & \textbf{6.921} \\
\bottomrule
\end{tabular}
\end{table}

\begin{table}[h]
\centering
\caption{Statistics used to define motion regimes.}
\label{tab:motion_stats}
\begin{tabular}{lcc}
\toprule
Statistic & Low & High \\
\midrule
Body keypoints confidence & 0.737 & 0.663 \\
Max body kp moving dist (px) & 21.268 & 216.269 \\
Var body kp moving dist (px) & 6.098 & 34.022 \\
Head keypoints confidence & 0.9856 & 0.9840 \\
Max head kp moving dist (px) & 2.258 & 34.201 \\
Var head kp moving dist (px) & 2.961 & 216.383 \\
\bottomrule
\end{tabular}
\end{table}

\subsection{Effect of Mask Cache}
Using the cached mask from step $t{-}1$ to gate audio injection at step $t$ markedly improves lip sync (Table~\ref{tab:mask_cache}; \textbf{6.921} vs 11.046 Sync-D). Without caching, identity assignment degrades toward a single-person behavior, confirming the necessity of cross-timestep spatial guidance.

\begin{table}[H]
\centering
\caption{Mask cache significantly improves multi-person lip sync.}
\begin{tabular}{lc}
\toprule
Condition & Sync-D$\downarrow$ \\
\midrule
With mask cache & \textbf{6.921} \\
Without mask cache & 11.046 \\
\bottomrule
\end{tabular}
\label{tab:mask_cache}
\end{table}

\subsection{Scaling Beyond Three People}
Table~\ref{tab:more_people} shows stable scaling to 4–5 speakers: Sync-D improves slightly (\textbf{6.608}) and AES increases (\textbf{3.992}) with a marginal IQA change. This aligns with our claim that per-identity mask prediction is independent, enabling more entities without collapsing lip-sync quality.

\begin{table}[H]
\centering
\caption{Stable performance when scaling to 4–5 subjects.}
\begin{tabular}{lccc}
\toprule
\#People & Sync-D$\downarrow$ & AES$\uparrow$ & IQA$\uparrow$ \\
\midrule
$\leq$3  & 6.670 & 3.467 & \textbf{4.757} \\
4--5     & \textbf{6.608} & \textbf{3.992} & 4.738 \\
\bottomrule
\end{tabular}
\label{tab:more_people}
\end{table}

\section{Algorithm of Our Model Implementation}
Please refer to Algorithm~\ref{alg:interacthuman} for our model implementation.
\begin{algorithm}[ht]
\caption{InterActHuman: Audio-Driven Multi-Concept Video Generation Inference}
\label{alg:interacthuman}
\begin{algorithmic}[1]
    \Require 
        \Statex Text prompt $T$, reference images $\{X_i\}_{i=1}^N$, identity-level audio $\{Y_i\}_{i=1}^N$
        \Statex Total diffusion steps $S$, mask injection threshold step $S_{\text{mask}}$
        \Statex VAE Encoder/Decoder, diffusion transformer
    \Ensure Generated video $V$
    
    \State \textbf{Preprocessing:}
    \For{$i \gets 1$ \textbf{to} $N$}
        \State $T_i \gets \text{Rephrase}(X_i, T)$ \Comment{Generate detailed prompt via Qwen2.5-VL.}
        \State $\mathbf{x}_i \gets \text{VAE\_Encoder}(X_i)$ \Comment{Encode reference image.}
        \State $\mathbf{a}_i \gets \text{wav2vec}(Y_i)$ \Comment{Extract audio features.}
    \EndFor
    \State $c_{\text{text}} \gets \{T, T_1, \ldots, T_N\}$ \Comment{Aggregate text conditions.}
    \State $z_S \sim \mathcal{N}(0, I)$ \Comment{Initialize the noisy video latent.}
    \State Initialize mask cache: $\{m_i^\text{prev}\} \gets 0$
    
    \For{$k \gets S$ \textbf{downto} $1$}
        \For{each DiT block layer $l$ that we inject conditions}
            \For{$i \gets 1$ \textbf{to} $N$}
                \State \textbf{Reference Injection:} Inject $\{\mathbf{x}_i\}$ via concatenation and self-attention.
                \State Compute hidden video feature $\mathbf{h}^{v}$ and reference feature $\mathbf{h}^{r}_i$.
                \State $\mathbf{Q}^{v} \gets \text{Proj}_q(\mathbf{h}^{v})$. \quad  $[\mathbf{K}^{r}_i,\, \mathbf{V}^{r}_i] \gets \text{Proj}_{k,v}(\mathbf{h}^{r}_i)$.
                \State Apply LayerNorm and 3D RoPE to the features.
                \State Compute cross-attention:
                \[
                \mathbf{p}^{(l)}_i \gets \operatorname{softmax}\Big(\frac{\mathbf{Q}^{v} (\mathbf{K}^{r}_i)^{\top}}{\sqrt{d}}\Big)\mathbf{V}^{r}_i.
                \]
                \State Predict layer mask: 
                \[
                m^{(l)}_i \gets \operatorname{sigmoid}\Big(\text{MLP}(\mathbf{p}^{(l)}_i)\Big).
                \]
            \EndFor
        \EndFor
        \State \textbf{Aggregate Masks:} 
        \[
        m_i \gets \frac{1}{L}\sum_{l=1}^{L} m^{(l)}_i \quad \forall~ i \in \{1,\ldots,N\}.
        \]
        \State Cache current masks: $\{m_i^\text{prev}\} \gets \{m_i\}$.
        
        \If{$k < S_{\text{mask}}$}
            \For{$i \gets 1$ \textbf{to} $N$}
                \State \textbf{Audio Injection}: inject local audio by updating the noisy latent $\mathbf{h}^{v}$.
                \State $\mathbf{Q}^{v} \gets \text{Proj}_q(\mathbf{h}^{v})$. \quad $[\mathbf{K}^{\text{mute}}_i,\, \mathbf{V}^{\text{mute}}_i] \gets \text{Proj}_{k,v}(\mathbf{a}_i^{\text{mute}})$. \quad $[\mathbf{K}_i,\, \mathbf{V}_i] \gets \text{Proj}_{k,v}(\mathbf{a}_i)$.
                \State Compute cross-attention:
                \[
                \mathbf{p}_i \gets \operatorname{softmax}\Big(\frac{\mathbf{Q}^{v} (\mathbf{K}_i)^{\top}}{\sqrt{d}}\Big)\mathbf{V}_i,  \quad \quad \mathbf{p}^{\text{mute}}_i \gets \operatorname{softmax}\Big(\frac{\mathbf{Q}^{v} (\mathbf{K}^{\text{mute}}_i)^{\top}}{\sqrt{d}}\Big)\mathbf{V}^{\text{mute}}_i, 
                \]
                \State Bind audio conditions:
                \[
                \mathbf{h}^{v} \gets \mathbf{h}^{v} + m_i \odot \mathbf{p}_i + (1-m_i) \odot \mathbf{p}_i^{\text{mute}},
                \]
                where $\odot$ denotes element-wise multiplication.
            \EndFor
        \EndIf
        
        \State Update latent via diffusion step via flow-matching formulas or custom samplers.
    \EndFor
    
    \State \textbf{Decoding:} Decode the final latent via VAE:
    \[
    V \gets \text{VAE\_Decoder}(z_0).
    \]
    \Return $V$
\end{algorithmic}
\end{algorithm}

\section{More Details of Our Model}

\subsection{Long Video Generation and Frame-Alignment Details}
We follow sliding-window strategies~\citep{jiang2024loopy,emo} by reusing several motion frames from the tail of window $w$ as the head of window $w\!+\!1$. Reference injection is unchanged; starting frames are optional for window 1.

\paragraph{Frame alignment flag.}
Per-frame masks have confidence scores from SAM2; frames with confidence $<0.5$ are excluded from the mask loss (flow-matching still applies). Videos are retained if other frames have reliable masks, so supervision remains temporally dense.

\subsection{Loss Design and Audio Stack Clarifications}
\paragraph{Focal loss.}
We adopt focal loss with $\alpha=0.25,\ \gamma=2$ to mitigate class imbalance (background vs person) and stabilize early training; BCE converges more slowly and is less stable initially.

\paragraph{Audio features.}
Audio tokens come from wav2vec~2.0; we do not add extra transformer blocks before injection. Spatial attention is unrestricted; temporally each latent attends to a local $\pm5$ token window.

\subsection{Implementation Details}
Our model was trained for 10,000 steps on 32 A800 GPUs with a learning rate of 3e-5. We adopted the PyTorch framework combined with Fully Sharded Data Parallel (FSDP) to finetune the DiT model across multiple GPUs. Specifically, the model was partitioned such that different GPUs handled distinct portions of the model’s parameters. We configured the effective batch size so that every node—comprising 8 GPUs—processed 2 videos simultaneously. With 32 GPUs in total (i.e., 4 nodes), this resulted in an overall effective batch size of 8 videos.
\subsection{Ablation Details}
In Tab. 4 and Fig. 5 of the paper, we validate our local audio injection designs via ablation on three variants: \emph{1) Global audio}~\citep{lin2025omnihuman} applies audio across the entire feature map;
\emph{2) ID Embedding} injects audio features with a learnable ID embedding without mask prediction;
\emph{3) Fixed Mask} uses predefined static spatial masks.

For global audio, it is identical to the pretrained audio-driven model~\citep{lin2025omnihuman}. For fixed mask, it is a manually input rectangle mask upon the audio cross-attention of pretrained audio-driven model~\citep{lin2025omnihuman}. It requires the generated human being static and it needs the user to ensure there is only one person inside a mask.

For ID embedding, here we provide some implementation details. Inspired by detection transformers~\citep{carion2020end}, we introduce a set of $N$ learnable ID embeddings $\mathbf{E}_{query} \in \mathbb{R}^{N \times C}$. These embeddings serve as identity tokens for the $N$ individuals. Specifically, we add the same learnable ID embedding $\mathbf{e}_{query} \in \mathbb{R}^{C}$ to a paired reference image and audio segment. In this way, we expect the model to implicitly match an individual (defined by a reference image) and audio segments in the output video to get synchronized lip movements. For the ID embedding dataset preparation, we incorporate multi-person videos into training. We use Active Speaker Detection (ASD) to extract talking persons' identities as a bounding box with identity-number. Then we use the bounding box and SAM2 masks to match the identity-number of each timestep of audio based on their IoU. Finally we use paired audio with timestep and reference image to generate video, and use flow-matching loss only to supervised it. No mask information is provided to the model and no mask supervision is adopted.

In the main paper, our experimental results in ablation study indicate that implicit matching of individuals and audio segments is worse than explicit matching with layout-aligned mask prediction and audio injection. It is worth noting that although our implementation shows this evidence, implicit matching is not necessarily worse than explicit matching. We show that a straight-forward implementation of implicit matching cannot generate satisfactory results, yet there could be better implementations to improve this result in the future.

\section{Details of Audio-driven Base Model's Architecture and Training}
\label{sec:omnihuman_details}
This appendix provides a detailed description of the network structure and training specifics for the pretrained audio-driven single-person video generation model, OmniHuman~\citep{lin2025omnihuman}. As noted in our main paper, the text-to-video base model undergoes post-pretraining for audio conditioning, following the OmniHuman~\citep{lin2025omnihuman} methodology. Subsequently, our multi-concept framework is built and trained on this foundation.

\subsection{Audio-driven Base Model}
Our foundational model is composed of a Variational Autoencoder (VAE) and a Latent Diffusion Transformer (DiT). The VAE includes an encoder, which compresses raw pixel data into a compact latent representation, and a decoder, which reconstructs the original pixel inputs from these latent features. The VAE achieves compression ratios of (4, 8, 8) for the (t, h, w) dimensions, respectively. Both the encoder and decoder utilize a temporally causal convolutional architecture, facilitating image and video compression across both spatial and temporal domains within the joint latent space. The denoising latents possess 16 channels.

The DiT Blocks are based on the dual-stream Diffusion Transformer (DiT)~\citep{esser2024sd3}. This transformer processes video and text tokens through multiple self-attention layers and feedforward networks (FFNs) to learn representations that are both shared and modality-specific. SwiGLU is employed as the activation function to enhance nonlinear modeling capabilities. Additionally, AdaSingle~\citep{chen2023pixart} is used for efficient timestep modulation. Moreover, two-thirds of the FFN weights in the deeper layers share parameters, creating a hybrid-stream design that preserves model capacity while substantially improving parameter efficiency.

\subsection{Mixed Training Strategy}

We utilize a multi-condition training strategy based on the framework established in OmniHuman~\citep{lin2025omnihuman}. Our method employs a two-stage progressive training scheme to develop the base model's capabilities. In the initial stage, the model is trained solely on text-to-video (T2V) data to build fundamental video generation abilities. The second stage introduces audio-synchronized datasets to expand the model's functionality to include audio-driven generation and reference image injection. For this second stage, we follow two core principles from OmniHuman: 1) Tasks with stronger conditioning can leverage tasks with weaker conditioning and their associated data to broaden the effective training dataset; 2) Tasks with stronger conditioning should be assigned proportionally lower training ratios. Guided by these principles, we first train the reference image injection capability before introducing the audio-driven generation objective. This staged methodology facilitates efficient knowledge transfer while ensuring stable training dynamics throughout the multi-condition learning process.

\section{Additional Details for Audio-driven Base Model Dataset}

\subsection{Dataset Curation}
We initially filter the T2V data using rules 1 through 6. Following this, we apply rule 7 to further refine the audio-driven data.

\textbf{1. Video Clip}. We begin by using PySceneDetect (~\href{https://github.com/Breakthrough/PySceneDetect}{https://github.com/Breakthrough/PySceneDetect}) to identify and trim shot transitions and fades within video clips. After this process, all clips are standardized to a duration of 5 to 30 seconds.

\textbf{2. Human}. Utilizing the annotated video captions, we implement a rule-based system to detect keywords such as "people", "human", "men", "women", "girl", and "boy". If any of these keywords are found, the video is categorized as human-related. This technique ensures the dataset is effectively filtered to comprise only videos pertinent to human activities and interactions.

\textbf{3. Subtitles}. We use PaddleOCR ( ~\href{https://github.com/PaddlePaddle/PaddleOCR}{https://github.com/PaddlePaddle/PaddleOCR}) to detect subtitles in the videos and remove clips where subtitles change. This step guarantees that the dataset emphasizes continuous and consistent visual content, reducing distractions from textual variations and improving the data quality for subsequent tasks.

\textbf{4. Visual Quality}. We utilize Q-align \citep{wu2023q} to evaluate the visual quality of the videos, filtering out clips that do not meet a predefined threshold. This procedure ensures the dataset maintains a high standard of visual clarity, which is essential for accurate analysis and robust model performance. By eliminating low-quality segments, we enhance the overall dependability and utility of the dataset.

\textbf{5. Aesthetics}. We employ Q-align \citep{wu2023q} to assess the aesthetic appeal of the videos and discard clips falling below a set threshold. Through aesthetic quality assessment, we can filter out videos containing post-production elements, thereby improving the quality of the training dataset. This measure ensures the dataset is composed of natural and unaltered video content, which better represents real-world scenarios and boosts the robustness and generalization of the trained models.

\textbf{6. Motion}. We use Raft~\citep{teed2020raft} to calculate the optical flow of the videos and filter out clips exhibiting overly intense motion. This step ensures that the dataset includes only video clips with moderate and significant motion, which are more appropriate for analysis and model training. By removing clips with extreme motion, we enhance the stability and quality of the dataset, leading to more dependable results.

\textbf{7. Syncnet}. For audio-driven content, we employ SyncNet \citep{syncnet} to determine if the lip movements are synchronized with the audio. Videos displaying considerable asynchronization are excluded. This step ensures that the dataset consists only of high-quality, synchronized audio-visual data. By removing out-of-sync segments, we improve the overall quality and reliability of the dataset. Ultimately, we gather 2,000 hours of audio-driven data.

\subsection{Dataset Analysis}
To gain a better understanding of the dataset's distribution, we perform an analysis across three dimensions. Detailed definitions for each dimension are provided below.

\begin{itemize}
\item \textbf{Human Size}. Human size indicates the portion of the human body visible within the video frame. It is categorized into the following levels: \textit{Portrait} (head and shoulders), \textit{Chest} (from head to chest), \textit{Waist} (from head to waist), \textit{Knees} (from head to knees), and \textit{Full Body} (the entire body is visible). To determine this, we first detect body keypoints using RTMpose~\citep{jiang2023rtmpose}, and then classify the human size based on the confidence scores of these keypoints. This method ensures a robust and accurate categorization of the visible human body extent in each video.
\item \textbf{Motion Amplitude}. After extracting body keypoints from the video, the amplitude of human motion is computed by measuring the displacement of the chest keypoint over time, relative to the width of the shoulders. Based on these calculations, we classify motion amplitude into four categories: \textit{Slight}($<0.1$), \textit{Moderate}($0.1-0.2$), \textit{Significant}($0.2-0.3$), and \textit{Extreme}($>0.3$).
\item \textbf{Scene}. Using video captioning, we employ Doubao to categorize videos into the following scenarios: \textit{Household}, \textit{Work}, \textit{Show}, \textit{Outdoor Adventure}, \textit{transportation}, \textit{Arts and Crafts}, \textit{Sports}, and \textit{Others}.
\item \textbf{Language}. For data containing audio, we use a language detection tool (~\href{https://github.com/Mimino666/langdetect}{https://github.com/Mimino666/langdetect}) to identify and count the types of languages present. These are categorized as \textit{English}, \textit{Chinese}, \textit{Spanish}, \textit{French}, \textit{German}, \textit{Urdu}, \textit{Hindi}, and \textit{Others}.
\end{itemize}

\section{User Study Details}
Here we show our questionnaires used in the user study.
\subsection{Multi-person Audio-Driven Video Generation Questionnaire}

Please take 10 minutes to complete the following ranking questions. For each question, consider the priority of the three videos based on the following dimensions and rank them accordingly:

(1) Lip-sync Accuracy: Based on the audio you hear, do you think the lip movements are accurate? Ideally, only one person's lip movements should correspond to the audio, but it doesn't matter specifically who is speaking. It is considered poor if multiple people are speaking simultaneously or if no one is speaking when there is audio. When multiple methods all correctly show only one person speaking, rank them based on the quality of lip-sync with the audio. Perfect synchronization is good; missing syllables or incorrect lip shapes for certain syllables is poor.

(2) Sense of Dialogue Among Multiple People: Is there a feeling of conversation between the individuals? It is considered good if it portrays a natural scenario where one person speaks and another listens. The judgment here is based on whether their expressions during the interaction appear natural and whether the overall feeling of the conversation is natural.

(3) Video Quality: If all the above criteria are tied, then prioritize the video with higher quality.

The priority of these three criteria decreases in the order listed. Aspect ratio and resolution should not be taken into account; you should not favor a particular aspect ratio or higher resolution, but rather focus on the inherent video quality itself.

\subsection{Multi-concept Video Generation Questionnaire}

Please take 8 minutes to complete the following ranking questions. For each question, consider the priority of the five videos from the perspective of consistency between the reference images and the appearance in the video, and rank them accordingly:

(1) Consistency with Reference Images: Based on the reference images, do you think all the listed reference images appear in the video? And does the appearance of those in the video match the images? Consider the priority in the following order from high to low:

Ranked highest: Appearance is consistent, and all reference images appear.

Next: All reference images appear, but the appearance of some reference images is inconsistent.

Ranked lowest: Some reference images do not appear.

(2) Video Quality: If the above criteria are tied, then prioritize the video with higher quality.

(3) Motion Strength: If all the above criteria are tied, then prioritize the video with a larger motion strength.

The priority of these criteria decreases in the order listed. Aspect ratio and resolution should not be taken into account; you should not favor a particular aspect ratio or higher resolution, but rather focus on the inherent video quality itself.
\end{document}

%% file: math_commands.tex
\usepackage{amsmath,amsfonts,bm}

\def\eqref#1{equation~\ref{#1}}

\def\1{\bm{1}}

\DeclareMathAlphabet{\mathsfit}{\encodingdefault}{\sfdefault}{m}{sl}
\SetMathAlphabet{\mathsfit}{bold}{\encodingdefault}{\sfdefault}{bx}{n}













%% file: iclr2026_conference.bbl
\begin{thebibliography}{102}
\providecommand{\natexlab}[1]{#1}
\providecommand{\url}[1]{\texttt{#1}}
\expandafter\ifx\csname urlstyle\endcsname\relax
  \providecommand{\doi}[1]{doi: #1}\else
  \providecommand{\doi}{doi: \begingroup \urlstyle{rm}\Url}\fi

\bibitem[pik()]{pikaart2024}
Pika art.
\newblock \url{https://pika.art/}.
\newblock Accessed: 2025-05-12.

\bibitem[Baevski et~al.(2020)Baevski, Zhou, Mohamed, and Auli]{baevski2020wav2vec2}
Alexei Baevski, Yuhao Zhou, Abdelrahman Mohamed, and Michael Auli.
\newblock wav2vec 2.0: A framework for self-supervised learning of speech representations.
\newblock \emph{Advances in neural information processing systems}, 33:\penalty0 12449--12460, 2020.

\bibitem[Bai et~al.(2025)Bai, Chen, Liu, Wang, Ge, Song, Dang, Wang, Wang, Tang, et~al.]{bai2025qwen2}
Shuai Bai, Keqin Chen, Xuejing Liu, Jialin Wang, Wenbin Ge, Sibo Song, Kai Dang, Peng Wang, Shijie Wang, Jun Tang, et~al.
\newblock Qwen2. 5-vl technical report.
\newblock \emph{arXiv preprint arXiv:2502.13923}, 2025.

\bibitem[Bain et~al.(2021)Bain, Nagrani, Varol, and Zisserman]{bain2021frozen}
Max Bain, Arsha Nagrani, G{\"u}l Varol, and Andrew Zisserman.
\newblock Frozen in time: A joint video and image encoder for end-to-end retrieval.
\newblock In \emph{Proceedings of the IEEE/CVF international conference on computer vision}, pp.\  1728--1738, 2021.

\bibitem[Bao et~al.(2024)Bao, Xiang, Yue, He, Zhu, Zheng, Zhao, Liu, Wang, and Zhu]{bao2024vidu}
Fan Bao, Chendong Xiang, Gang Yue, Guande He, Hongzhou Zhu, Kaiwen Zheng, Min Zhao, Shilong Liu, Yaole Wang, and Jun Zhu.
\newblock Vidu: a highly consistent, dynamic and skilled text-to-video generator with diffusion models.
\newblock \emph{arXiv preprint arXiv:2405.04233}, 2024.

\bibitem[Bar-Tal et~al.(2024)Bar-Tal, Chefer, Tov, Herrmann, Paiss, Zada, Ephrat, Hur, Li, Michaeli, et~al.]{bar2024lumiere}
Omer Bar-Tal, Hila Chefer, Omer Tov, Charles Herrmann, Roni Paiss, Shiran Zada, Ariel Ephrat, Junhwa Hur, Yuanzhen Li, Tomer Michaeli, et~al.
\newblock Lumiere: A space-time diffusion model for video generation.
\newblock \emph{arXiv preprint arXiv:2401.12945}, 2024.

\bibitem[Blattmann et~al.(2023{\natexlab{a}})Blattmann, Dockhorn, Kulal, Mendelevitch, Kilian, Lorenz, Levi, English, Voleti, Letts, et~al.]{svd}
Andreas Blattmann, Tim Dockhorn, Sumith Kulal, Daniel Mendelevitch, Maciej Kilian, Dominik Lorenz, Yam Levi, Zion English, Vikram Voleti, Adam Letts, et~al.
\newblock Stable video diffusion: Scaling latent video diffusion models to large datasets.
\newblock \emph{arXiv preprint arXiv:2311.15127}, 2023{\natexlab{a}}.

\bibitem[Blattmann et~al.(2023{\natexlab{b}})Blattmann, Rombach, Ling, Dockhorn, Kim, Fidler, and Kreis]{ayl}
Andreas Blattmann, Robin Rombach, Huan Ling, Tim Dockhorn, Seung~Wook Kim, Sanja Fidler, and Karsten Kreis.
\newblock Align your latents: High-resolution video synthesis with latent diffusion models.
\newblock In \emph{Proceedings of the IEEE/CVF Conference on Computer Vision and Pattern Recognition}, pp.\  22563--22575, 2023{\natexlab{b}}.

\bibitem[Brooks et~al.(2022)Brooks, Hellsten, Aittala, Wang, Aila, Lehtinen, Liu, Efros, and Karras]{videogan}
Tim Brooks, Janne Hellsten, Miika Aittala, Ting-Chun Wang, Timo Aila, Jaakko Lehtinen, Ming-Yu Liu, Alexei Efros, and Tero Karras.
\newblock Generating long videos of dynamic scenes.
\newblock \emph{Advances in Neural Information Processing Systems}, 35:\penalty0 31769--31781, 2022.

\bibitem[Brooks et~al.(2024)Brooks, Peebles, Holmes, DePue, Guo, Jing, Schnurr, Taylor, Luhman, Luhman, Ng, Wang, and Ramesh]{videoworldsimulators2024}
Tim Brooks, Bill Peebles, Connor Holmes, Will DePue, Yufei Guo, Li~Jing, David Schnurr, Joe Taylor, Troy Luhman, Eric Luhman, Clarence Ng, Ricky Wang, and Aditya Ramesh.
\newblock Video generation models as world simulators.
\newblock 2024.
\newblock URL \url{https://openai.com/research/video-generation-models-as-world-simulators}.

\bibitem[Carion et~al.(2020)Carion, Massa, Synnaeve, Usunier, Kirillov, and Zagoruyko]{carion2020end}
Nicolas Carion, Francisco Massa, Gabriel Synnaeve, Nicolas Usunier, Alexander Kirillov, and Sergey Zagoruyko.
\newblock End-to-end object detection with transformers.
\newblock In \emph{European conference on computer vision}, pp.\  213--229. Springer, 2020.

\bibitem[Chen et~al.(2023)Chen, Yu, Ge, Yao, Xie, Wu, Wang, Kwok, Luo, Lu, et~al.]{chen2023pixart}
Junsong Chen, Jincheng Yu, Chongjian Ge, Lewei Yao, Enze Xie, Yue Wu, Zhongdao Wang, James Kwok, Ping Luo, Huchuan Lu, et~al.
\newblock Pixart-$\alpha$: Fast training of diffusion transformer for photorealistic text-to-image synthesis.
\newblock \emph{arXiv preprint arXiv:2310.00426}, 2023.

\bibitem[Chen et~al.(2025{\natexlab{a}})Chen, Siarohin, Menapace, Fang, Lee, Skorokhodov, Aberman, Zhu, Yang, and Tulyakov]{chen2025multi}
Tsai-Shien Chen, Aliaksandr Siarohin, Willi Menapace, Yuwei Fang, Kwot~Sin Lee, Ivan Skorokhodov, Kfir Aberman, Jun-Yan Zhu, Ming-Hsuan Yang, and Sergey Tulyakov.
\newblock Multi-subject open-set personalization in video generation.
\newblock \emph{arXiv preprint arXiv:2501.06187}, 2025{\natexlab{a}}.

\bibitem[Chen et~al.(2025{\natexlab{b}})Chen, Liang, Zhou, Huang, Ma, Tang, Lin, Zhou, and Lu]{chen2025hunyuanvideo}
Yi~Chen, Sen Liang, Zixiang Zhou, Ziyao Huang, Yifeng Ma, Junshu Tang, Qin Lin, Yuan Zhou, and Qinglin Lu.
\newblock Hunyuanvideo-avatar: High-fidelity audio-driven human animation for multiple characters.
\newblock \emph{arXiv preprint arXiv:2505.20156}, 2025{\natexlab{b}}.

\bibitem[Chen et~al.(2024)Chen, Cao, Chen, Li, and Ma]{chen2024echomimic}
Zhiyuan Chen, Jiajiong Cao, Zhiquan Chen, Yuming Li, and Chenguang Ma.
\newblock Echomimic: Lifelike audio-driven portrait animations through editable landmark conditions.
\newblock \emph{arXiv preprint arXiv:2407.08136}, 2024.

\bibitem[Chung \& Zisserman(2017)Chung and Zisserman]{syncnet}
Joon~Son Chung and Andrew Zisserman.
\newblock Out of time: automated lip sync in the wild.
\newblock In \emph{Computer Vision--ACCV 2016 Workshops: ACCV 2016 International Workshops, Taipei, Taiwan, November 20-24, 2016, Revised Selected Papers, Part II 13}, pp.\  251--263. Springer, 2017.

\bibitem[Corona et~al.(2024)Corona, Zanfir, Bazavan, Kolotouros, Alldieck, and Sminchisescu]{VLogger}
Enric Corona, Andrei Zanfir, Eduard~Gabriel Bazavan, Nikos Kolotouros, Thiemo Alldieck, and Cristian Sminchisescu.
\newblock Vlogger: Multimodal diffusion for embodied avatar synthesis.
\newblock \emph{arXiv preprint arXiv:2403.08764}, 2024.

\bibitem[Cui et~al.(2024)Cui, Li, Zhan, Shang, Cheng, Ma, Mu, Zhou, Wang, and Zhu]{hallo3}
Jiahao Cui, Hui Li, Yun Zhan, Hanlin Shang, Kaihui Cheng, Yuqi Ma, Shan Mu, Hang Zhou, Jingdong Wang, and Siyu Zhu.
\newblock Hallo3: Highly dynamic and realistic portrait image animation with diffusion transformer networks.
\newblock \emph{arXiv preprint arXiv:2412.00733}, 2024.

\bibitem[Deng et~al.(2019)Deng, Guo, Xue, and Zafeiriou]{deng2019arcface}
Jiankang Deng, Jia Guo, Niannan Xue, and Stefanos Zafeiriou.
\newblock Arcface: Additive angular margin loss for deep face recognition.
\newblock In \emph{CVPR}, pp.\  4690--4699, 2019.

\bibitem[Esser et~al.(2024)Esser, Kulal, Blattmann, Entezari, M{\"u}ller, Saini, Levi, Lorenz, Sauer, Boesel, et~al.]{esser2024sd3}
Patrick Esser, Sumith Kulal, Andreas Blattmann, Rahim Entezari, Jonas M{\"u}ller, Harry Saini, Yam Levi, Dominik Lorenz, Axel Sauer, Frederic Boesel, et~al.
\newblock Scaling rectified flow transformers for high-resolution image synthesis.
\newblock In \emph{Forty-first International Conference on Machine Learning}, 2024.

\bibitem[Fei et~al.(2025{\natexlab{a}})Fei, Li, Qiu, Wang, Dou, Wang, Xu, Fan, Chen, Li, et~al.]{fei2025skyreels}
Zhengcong Fei, Debang Li, Di~Qiu, Jiahua Wang, Yikun Dou, Rui Wang, Jingtao Xu, Mingyuan Fan, Guibin Chen, Yang Li, et~al.
\newblock Skyreels-a2: Compose anything in video diffusion transformers.
\newblock \emph{arXiv preprint arXiv:2504.02436}, 2025{\natexlab{a}}.

\bibitem[Fei et~al.(2025{\natexlab{b}})Fei, Li, Qiu, Yu, and Fan]{fei2025ingredients}
Zhengcong Fei, Debang Li, Di~Qiu, Changqian Yu, and Mingyuan Fan.
\newblock Ingredients: Blending custom photos with video diffusion transformers.
\newblock \emph{arXiv preprint arXiv:2501.01790}, 2025{\natexlab{b}}.

\bibitem[Feng et~al.(2025)Feng, Liu, Liu, Wang, Vahdat, and Nie]{feng2025blobgen}
Weixi Feng, Chao Liu, Sifei Liu, William~Yang Wang, Arash Vahdat, and Weili Nie.
\newblock Blobgen-vid: Compositional text-to-video generation with blob video representations.
\newblock \emph{arXiv preprint arXiv:2501.07647}, 2025.

\bibitem[Guo et~al.(2023)Guo, Yang, Rao, Liang, Wang, Qiao, Agrawala, Lin, and Dai]{guo2024animatediff}
Yuwei Guo, Ceyuan Yang, Anyi Rao, Zhengyang Liang, Yaohui Wang, Yu~Qiao, Maneesh Agrawala, Dahua Lin, and Bo~Dai.
\newblock Animatediff: Animate your personalized text-to-image diffusion models without specific tuning.
\newblock In \emph{International Conference on Learning Representations (ICLR)}, 2023.

\bibitem[Gupta et~al.(2023)Gupta, Yu, Sohn, Gu, Hahn, Fei-Fei, Essa, Jiang, and Lezama]{walt}
Agrim Gupta, Lijun Yu, Kihyuk Sohn, Xiuye Gu, Meera Hahn, Li~Fei-Fei, Irfan Essa, Lu~Jiang, and Jos{\'e} Lezama.
\newblock Photorealistic video generation with diffusion models.
\newblock \emph{arXiv preprint arXiv:2312.06662}, 2023.

\bibitem[He et~al.(2023)He, Guo, Yu, Wang, Zhu, An, Li, Tan, Wang, Hu, et~al.]{he2023gaia}
Tianyu He, Junliang Guo, Runyi Yu, Yuchi Wang, Jialiang Zhu, Kaikai An, Leyi Li, Xu~Tan, Chunyu Wang, Han Hu, et~al.
\newblock Gaia: Zero-shot talking avatar generation.
\newblock \emph{arXiv preprint arXiv:2311.15230}, 2023.

\bibitem[He et~al.(2024)He, Liu, Qian, Wang, Hu, Cao, Yan, Zhou, and Zhang]{he2024id}
Xuanhua He, Quande Liu, Shengju Qian, Xin Wang, Tao Hu, Ke~Cao, Keyu Yan, Man Zhou, and Jie Zhang.
\newblock Id-animator: Zero-shot identity-preserving human video generation.
\newblock \emph{arXiv preprint arXiv:2404.15275}, 2024.

\bibitem[Ho et~al.(2020)Ho, Jain, and Abbeel]{jonathan2020ddpm}
Jonathan Ho, Ajay Jain, and Pieter Abbeel.
\newblock Denoising diffusion probabilistic models.
\newblock In H.~Larochelle, M.~Ranzato, R.~Hadsell, M.F. Balcan, and H.~Lin (eds.), \emph{Advances in Neural Information Processing Systems}, volume~33, pp.\  6840--6851. Curran Associates, Inc., 2020.
\newblock URL \url{https://proceedings.neurips.cc/paper/2020/file/4c5bcfec8584af0d967f1ab10179ca4b-Paper.pdf}.

\bibitem[Ho et~al.(2022)Ho, Salimans, Gritsenko, Chan, Norouzi, and Fleet]{vdm}
Jonathan Ho, Tim Salimans, Alexey Gritsenko, William Chan, Mohammad Norouzi, and David~J Fleet.
\newblock Video diffusion models.
\newblock \emph{Advances in Neural Information Processing Systems}, 35:\penalty0 8633--8646, 2022.

\bibitem[Hogue et~al.(2024)Hogue, Zhang, Daruger, Tian, and Guo]{diffted}
Steven Hogue, Chenxu Zhang, Hamza Daruger, Yapeng Tian, and Xiaohu Guo.
\newblock Diffted: One-shot audio-driven ted talk video generation with diffusion-based co-speech gestures.
\newblock In \emph{Proceedings of the IEEE/CVF Conference on Computer Vision and Pattern Recognition}, pp.\  1922--1931, 2024.

\bibitem[Hong et~al.(2022)Hong, Ding, Zheng, Liu, and Tang]{hong2022cogvideo}
Wenyi Hong, Ming Ding, Wendi Zheng, Xinghan Liu, and Jie Tang.
\newblock Cogvideo: Large-scale pretraining for text-to-video generation via transformers.
\newblock \emph{arXiv preprint arXiv:2205.15868}, 2022.

\bibitem[Hu(2024)]{aa}
Li~Hu.
\newblock Animate anyone: Consistent and controllable image-to-video synthesis for character animation.
\newblock In \emph{Proceedings of the IEEE/CVF Conference on Computer Vision and Pattern Recognition}, pp.\  8153--8163, 2024.

\bibitem[Huang et~al.(2020)Huang, Wang, Tai, Liu, Shen, Li, Li, and Huang]{huang2020curricularface}
Yuge Huang, Yuhan Wang, Ying Tai, Xiaoming Liu, Pengcheng Shen, Shaoxin Li, Jilin Li, and Feiyue Huang.
\newblock Curricularface: adaptive curriculum learning loss for deep face recognition.
\newblock In \emph{proceedings of the IEEE/CVF conference on computer vision and pattern recognition}, pp.\  5901--5910, 2020.

\bibitem[Huang et~al.(2025)Huang, Yuan, Liu, Wang, Wang, Zhang, Wan, Zhang, and Gai]{huang2025conceptmaster}
Yuzhou Huang, Ziyang Yuan, Quande Liu, Qiulin Wang, Xintao Wang, Ruimao Zhang, Pengfei Wan, Di~Zhang, and Kun Gai.
\newblock Conceptmaster: Multi-concept video customization on diffusion transformer models without test-time tuning.
\newblock \emph{arXiv preprint arXiv:2501.04698}, 2025.

\bibitem[Jiang et~al.(2024{\natexlab{a}})Jiang, Liang, Yang, Lin, Zhong, and Zheng]{jiang2024loopy}
Jianwen Jiang, Chao Liang, Jiaqi Yang, Gaojie Lin, Tianyun Zhong, and Yanbo Zheng.
\newblock Loopy: Taming audio-driven portrait avatar with long-term motion dependency.
\newblock \emph{arXiv preprint arXiv:2409.02634}, 2024{\natexlab{a}}.

\bibitem[Jiang et~al.(2024{\natexlab{b}})Jiang, Lin, Rong, Liang, Zhu, Yang, and Zhong]{jiang2024mobileportrait}
Jianwen Jiang, Gaojie Lin, Zhengkun Rong, Chao Liang, Yongming Zhu, Jiaqi Yang, and Tianyun Zhong.
\newblock Mobileportrait: Real-time one-shot neural head avatars on mobile devices.
\newblock \emph{arXiv preprint arXiv:2407.05712}, 2024{\natexlab{b}}.

\bibitem[Jiang et~al.(2023)Jiang, Lu, Zhang, Ma, Han, Lyu, Li, and Chen]{jiang2023rtmpose}
Tao Jiang, Peng Lu, Li~Zhang, Ningsheng Ma, Rui Han, Chengqi Lyu, Yining Li, and Kai Chen.
\newblock Rtmpose: Real-time multi-person pose estimation based on mmpose.
\newblock \emph{arXiv preprint arXiv:2303.07399}, 2023.

\bibitem[Jiang et~al.(2024{\natexlab{c}})Jiang, Wu, Yang, Si, Lin, Qiao, Loy, and Liu]{jiang2024videobooth}
Yuming Jiang, Tianxing Wu, Shuai Yang, Chenyang Si, Dahua Lin, Yu~Qiao, Chen~Change Loy, and Ziwei Liu.
\newblock Videobooth: Diffusion-based video generation with image prompts.
\newblock In \emph{Proceedings of the IEEE/CVF Conference on Computer Vision and Pattern Recognition}, pp.\  6689--6700, 2024{\natexlab{c}}.

\bibitem[Karras et~al.(2022)Karras, Aittala, Aila, and Laine]{karras2022edm}
Tero Karras, Miika Aittala, Timo Aila, and Samuli Laine.
\newblock Elucidating the design space of diffusion-based generative models.
\newblock \emph{Advances in neural information processing systems}, 35:\penalty0 26565--26577, 2022.

\bibitem[Kingma \& Welling(2013)Kingma and Welling]{kingma2013auto}
Diederik~P Kingma and Max Welling.
\newblock Auto-encoding variational bayes.
\newblock \emph{arXiv preprint arXiv:1312.6114}, 2013.

\bibitem[Kirillov et~al.(2023)Kirillov, Mintun, Ravi, Mao, Rolland, Gustafson, Xiao, Whitehead, Berg, Lo, et~al.]{kirillov2023segment}
Alexander Kirillov, Eric Mintun, Nikhila Ravi, Hanzi Mao, Chloe Rolland, Laura Gustafson, Tete Xiao, Spencer Whitehead, Alexander~C Berg, Wan-Yen Lo, et~al.
\newblock Segment anything.
\newblock In \emph{Proceedings of the IEEE/CVF International Conference on Computer Vision}, pp.\  4015--4026, 2023.

\bibitem[Kong et~al.(2024)Kong, Tian, Zhang, Min, Dai, Zhou, Xiong, Li, Wu, Zhang, et~al.]{kong2024hunyuanvideo}
Weijie Kong, Qi~Tian, Zijian Zhang, Rox Min, Zuozhuo Dai, Jin Zhou, Jiangfeng Xiong, Xin Li, Bo~Wu, Jianwei Zhang, et~al.
\newblock Hunyuanvideo: A systematic framework for large video generative models.
\newblock \emph{arXiv preprint arXiv:2412.03603}, 2024.

\bibitem[Kong et~al.(2025)Kong, Gao, Zhang, Kang, Wei, Cai, Chen, and Luo]{kong2025let}
Zhe Kong, Feng Gao, Yong Zhang, Zhuoliang Kang, Xiaoming Wei, Xunliang Cai, Guanying Chen, and Wenhan Luo.
\newblock Let them talk: Audio-driven multi-person conversational video generation.
\newblock \emph{arXiv preprint arXiv:2505.22647}, 2025.

\bibitem[{Kuaishou}(2024)]{kuaishou2024}
{Kuaishou}.
\newblock Kling video model.
\newblock \url{https://kling.kuaishou.com/en}, 2024.
\newblock Accessed: 29 April 2025.

\bibitem[Li et~al.(2024)Li, Xu, Zhan, Mu, Li, Cheng, Chen, Chen, Ye, Wang, et~al.]{li2024openhumanvid}
Hui Li, Mingwang Xu, Yun Zhan, Shan Mu, Jiaye Li, Kaihui Cheng, Yuxuan Chen, Tan Chen, Mao Ye, Jingdong Wang, et~al.
\newblock Openhumanvid: A large-scale high-quality dataset for enhancing human-centric video generation.
\newblock \emph{arXiv preprint arXiv:2412.00115}, 2024.

\bibitem[Li et~al.(2018)Li, Min, Shen, Carlson, and Carin]{text2video}
Yitong Li, Martin Min, Dinghan Shen, David Carlson, and Lawrence Carin.
\newblock Video generation from text.
\newblock In \emph{Proceedings of the AAAI conference on artificial intelligence}, volume~32, 2018.

\bibitem[Lin et~al.(2024)Lin, Jiang, Liang, Zhong, Yang, and Zheng]{lin2024cyberhost}
Gaojie Lin, Jianwen Jiang, Chao Liang, Tianyun Zhong, Jiaqi Yang, and Yanbo Zheng.
\newblock Cyberhost: Taming audio-driven avatar diffusion model with region codebook attention.
\newblock \emph{arXiv preprint arXiv:2409.01876}, 2024.

\bibitem[Lin et~al.(2025{\natexlab{a}})Lin, Jiang, Yang, Zheng, and Liang]{lin2025omnihuman}
Gaojie Lin, Jianwen Jiang, Jiaqi Yang, Zerong Zheng, and Chao Liang.
\newblock Omnihuman-1: Rethinking the scaling-up of one-stage conditioned human animation models.
\newblock \emph{arXiv preprint arXiv:2502.01061}, 2025{\natexlab{a}}.

\bibitem[Lin et~al.(2025{\natexlab{b}})Lin, Xia, Ren, Yang, Xiao, and Jiang]{lin2025apt}
Shanchuan Lin, Xin Xia, Yuxi Ren, Ceyuan Yang, Xuefeng Xiao, and Lu~Jiang.
\newblock Diffusion adversarial post-training for one-step video generation.
\newblock \emph{arXiv preprint arXiv:2501.08316}, 2025{\natexlab{b}}.

\bibitem[Lin et~al.(2017)Lin, Goyal, Girshick, He, and Doll{\'a}r]{lin2017focal}
Tsung-Yi Lin, Priya Goyal, Ross Girshick, Kaiming He, and Piotr Doll{\'a}r.
\newblock Focal loss for dense object detection.
\newblock In \emph{Proceedings of the IEEE international conference on computer vision}, pp.\  2980--2988, 2017.

\bibitem[Lipman et~al.(2022)Lipman, Chen, Ben-Hamu, Nickel, and Le]{lipman2022flow}
Yaron Lipman, Ricky~TQ Chen, Heli Ben-Hamu, Maximilian Nickel, and Matt Le.
\newblock Flow matching for generative modeling.
\newblock \emph{arXiv preprint arXiv:2210.02747}, 2022.

\bibitem[Liu et~al.(2025)Liu, Ma, Li, Chen, Liu, He, and Wu]{liu2025phantom}
Lijie Liu, Tianxiang Ma, Bingchuan Li, Zhuowei Chen, Jiawei Liu, Qian He, and Xinglong Wu.
\newblock Phantom: Subject-consistent video generation via cross-modal alignment.
\newblock \emph{arXiv preprint arXiv:2502.11079}, 2025.

\bibitem[Liu et~al.(2023)Liu, Zeng, Ren, Li, Zhang, Yang, Jiang, Li, Yang, Su, et~al.]{liu2023grounding}
Shilong Liu, Zhaoyang Zeng, Tianhe Ren, Feng Li, Hao Zhang, Jie Yang, Qing Jiang, Chunyuan Li, Jianwei Yang, Hang Su, et~al.
\newblock Grounding dino: Marrying dino with grounded pre-training for open-set object detection.
\newblock \emph{arXiv preprint arXiv:2303.05499}, 2023.

\bibitem[Liu et~al.(2022)Liu, Gong, and Liu]{liu2022flow}
Xingchao Liu, Chengyue Gong, and Qiang Liu.
\newblock Flow straight and fast: Learning to generate and transfer data with rectified flow.
\newblock \emph{arXiv preprint arXiv:2209.03003}, 2022.

\bibitem[Meng et~al.(2024)Meng, Zhang, Li, and Ma]{EchomimicV2}
Rang Meng, Xingyu Zhang, Yuming Li, and Chenguang Ma.
\newblock Echomimicv2: Towards striking, simplified, and semi-body human animation.
\newblock \emph{arXiv preprint arXiv:2411.10061}, 2024.

\bibitem[Nagrani et~al.(2017)Nagrani, Chung, and Zisserman]{nagrani2017voxceleb}
A~Nagrani, J~Chung, and A~Zisserman.
\newblock Voxceleb: a large-scale speaker identification dataset.
\newblock \emph{Interspeech 2017}, 2017.

\bibitem[Oquab et~al.(2023)Oquab, Darcet, Moutakanni, Vo, Szafraniec, Khalidov, Fernandez, Haziza, Massa, El-Nouby, et~al.]{oquab2023dinov2}
Maxime Oquab, Timoth{\'e}e Darcet, Th{\'e}o Moutakanni, Huy Vo, Marc Szafraniec, Vasil Khalidov, Pierre Fernandez, Daniel Haziza, Francisco Massa, Alaaeldin El-Nouby, et~al.
\newblock Dinov2: Learning robust visual features without supervision.
\newblock \emph{arXiv preprint arXiv:2304.07193}, 2023.

\bibitem[Peebles \& Xie(2023)Peebles and Xie]{peebles2023scalable}
William Peebles and Saining Xie.
\newblock Scalable diffusion models with transformers.
\newblock In \emph{Proceedings of the IEEE/CVF International Conference on Computer Vision}, pp.\  4195--4205, 2023.

\bibitem[Polyak et~al.(2024)Polyak, Zohar, Brown, Tjandra, Sinha, Lee, Vyas, Shi, Ma, Chuang, et~al.]{polyak2024moviegen}
Adam Polyak, Amit Zohar, Andrew Brown, Andros Tjandra, Animesh Sinha, Ann Lee, Apoorv Vyas, Bowen Shi, Chih-Yao Ma, Ching-Yao Chuang, et~al.
\newblock Movie gen: A cast of media foundation models.
\newblock \emph{arXiv preprint arXiv:2410.13720}, 2024.

\bibitem[Qi et~al.(2023)Qi, Cun, Zhang, Lei, Wang, Shan, and Chen]{qi2023fatezero}
Chenyang Qi, Xiaodong Cun, Yong Zhang, Chenyang Lei, Xintao Wang, Ying Shan, and Qifeng Chen.
\newblock Fatezero: Fusing attentions for zero-shot text-based video editing.
\newblock \emph{arXiv:2303.09535}, 2023.

\bibitem[Radford et~al.(2021)Radford, Kim, Hallacy, Ramesh, Goh, Agarwal, Sastry, Askell, Mishkin, Clark, et~al.]{radford2021learning}
Alec Radford, Jong~Wook Kim, Chris Hallacy, Aditya Ramesh, Gabriel Goh, Sandhini Agarwal, Girish Sastry, Amanda Askell, Pamela Mishkin, Jack Clark, et~al.
\newblock Learning transferable visual models from natural language supervision.
\newblock In \emph{International conference on machine learning}, pp.\  8748--8763. PMLR, 2021.

\bibitem[Ren et~al.(2024)Ren, Liu, Zeng, Lin, Li, Cao, Chen, Huang, Chen, Yan, et~al.]{ren2024grounded}
Tianhe Ren, Shilong Liu, Ailing Zeng, Jing Lin, Kunchang Li, He~Cao, Jiayu Chen, Xinyu Huang, Yukang Chen, Feng Yan, et~al.
\newblock Grounded sam: Assembling open-world models for diverse visual tasks.
\newblock \emph{arXiv preprint arXiv:2401.14159}, 2024.

\bibitem[Seawead et~al.(2025)Seawead, Yang, Lin, Zhao, Lin, Ma, Guo, Chen, Qi, Wang, et~al.]{seawead2025seaweed}
Team Seawead, Ceyuan Yang, Zhijie Lin, Yang Zhao, Shanchuan Lin, Zhibei Ma, Haoyuan Guo, Hao Chen, Lu~Qi, Sen Wang, et~al.
\newblock Seaweed-7b: Cost-effective training of video generation foundation model.
\newblock \emph{arXiv preprint arXiv:2504.08685}, 2025.

\bibitem[Shao et~al.(2024)Shao, Pang, Zheng, Sun, and Liu]{shao2024human4dit}
Ruizhi Shao, Youxin Pang, Zerong Zheng, Jingxiang Sun, and Yebin Liu.
\newblock Human4dit: Free-view human video generation with 4d diffusion transformer.
\newblock \emph{arXiv preprint arXiv:2405.17405}, 2024.

\bibitem[Siarohin et~al.(2019)Siarohin, Lathuili{\`e}re, Tulyakov, Ricci, and Sebe]{siarohin2019fomm}
Aliaksandr Siarohin, St{\'e}phane Lathuili{\`e}re, Sergey Tulyakov, Elisa Ricci, and Nicu Sebe.
\newblock First order motion model for image animation.
\newblock \emph{Advances in neural information processing systems}, 32, 2019.

\bibitem[Siarohin et~al.(2021)Siarohin, Woodford, Ren, Chai, and Tulyakov]{siarohin2021mraa}
Aliaksandr Siarohin, Oliver~J Woodford, Jian Ren, Menglei Chai, and Sergey Tulyakov.
\newblock Motion representations for articulated animation.
\newblock In \emph{Proceedings of the IEEE/CVF Conference on Computer Vision and Pattern Recognition}, pp.\  13653--13662, 2021.

\bibitem[Singer et~al.(2022)Singer, Polyak, Hayes, Yin, An, Zhang, Hu, Yang, Ashual, Gafni, et~al.]{singer2022make}
Uriel Singer, Adam Polyak, Thomas Hayes, Xi~Yin, Jie An, Songyang Zhang, Qiyuan Hu, Harry Yang, Oron Ashual, Oran Gafni, et~al.
\newblock Make-a-video: Text-to-video generation without text-video data.
\newblock \emph{arXiv preprint arXiv:2209.14792}, 2022.

\bibitem[Song et~al.(2021)Song, Meng, and Ermon]{song2021ddim}
Jiaming Song, Chenlin Meng, and Stefano Ermon.
\newblock Denoising diffusion implicit models.
\newblock In \emph{International Conference on Learning Representations}, 2021.
\newblock URL \url{https://openreview.net/forum?id=St1giarCHLP}.

\bibitem[Song et~al.(2020)Song, Sohl-Dickstein, Kingma, Kumar, Ermon, and Poole]{song2020score}
Yang Song, Jascha Sohl-Dickstein, Diederik~P Kingma, Abhishek Kumar, Stefano Ermon, and Ben Poole.
\newblock Score-based generative modeling through stochastic differential equations.
\newblock \emph{arXiv preprint arXiv:2011.13456}, 2020.

\bibitem[Stypulkowski et~al.(2024)Stypulkowski, Vougioukas, He, Zieba, Petridis, and Pantic]{stypulkowski2024diffused}
Michal Stypulkowski, Konstantinos Vougioukas, Sen He, Maciej Zieba, Stavros Petridis, and Maja Pantic.
\newblock Diffused heads: Diffusion models beat gans on talking-face generation.
\newblock In \emph{Proceedings of the IEEE/CVF Winter Conference on Applications of Computer Vision}, pp.\  5091--5100, 2024.

\bibitem[Teed \& Deng(2020)Teed and Deng]{teed2020raft}
Zachary Teed and Jia Deng.
\newblock Raft: Recurrent all-pairs field transforms for optical flow.
\newblock In \emph{Computer Vision--ECCV 2020: 16th European Conference, Glasgow, UK, August 23--28, 2020, Proceedings, Part II 16}, pp.\  402--419. Springer, 2020.

\bibitem[Tian et~al.(2025{\natexlab{a}})Tian, Hu, Wang, Zhang, and Bo]{EMO2}
Linrui Tian, Siqi Hu, Qi~Wang, Bang Zhang, and Liefeng Bo.
\newblock Emo2: End-effector guided audio-driven avatar video generation.
\newblock \emph{arXiv preprint arXiv:2501.10687}, 2025{\natexlab{a}}.

\bibitem[Tian et~al.(2025{\natexlab{b}})Tian, Wang, Zhang, and Bo]{emo}
Linrui Tian, Qi~Wang, Bang Zhang, and Liefeng Bo.
\newblock Emo: Emote portrait alive generating expressive portrait videos with audio2video diffusion model under weak conditions.
\newblock In \emph{European Conference on Computer Vision}, pp.\  244--260. Springer, 2025{\natexlab{b}}.

\bibitem[Unterthiner et~al.()Unterthiner, van Steenkiste, Kurach, Marinier, Michalski, and Gelly]{unterthiner2019fvd}
Thomas Unterthiner, Sjoerd van Steenkiste, Karol Kurach, Rapha{\"e}l Marinier, Marcin Michalski, and Sylvain Gelly.
\newblock Fvd: A new metric for video generation.

\bibitem[Vaswani et~al.(2017)Vaswani, Shazeer, Parmar, Uszkoreit, Jones, Gomez, Kaiser, and Polosukhin]{vaswani2017attention}
Ashish Vaswani, Noam Shazeer, Niki Parmar, Jakob Uszkoreit, Llion Jones, Aidan~N Gomez, {\L}ukasz Kaiser, and Illia Polosukhin.
\newblock Attention is all you need.
\newblock \emph{Advances in neural information processing systems}, 30, 2017.

\bibitem[Villegas et~al.(2022)Villegas, Babaeizadeh, Kindermans, Moraldo, Zhang, Saffar, Castro, Kunze, and Erhan]{villegas2022phenaki}
Ruben Villegas, Mohammad Babaeizadeh, Pieter-Jan Kindermans, Hernan Moraldo, Han Zhang, Mohammad~Taghi Saffar, Santiago Castro, Julius Kunze, and Dumitru Erhan.
\newblock Phenaki: Variable length video generation from open domain textual descriptions.
\newblock In \emph{International Conference on Learning Representations}, 2022.

\bibitem[Wang et~al.(2025{\natexlab{a}})Wang, Ai, Wen, Mao, Xie, Chen, Yu, Zhao, Yang, Zeng, et~al.]{wang2025wan}
Ang Wang, Baole Ai, Bin Wen, Chaojie Mao, Chen-Wei Xie, Di~Chen, Feiwu Yu, Haiming Zhao, Jianxiao Yang, Jianyuan Zeng, et~al.
\newblock Wan: Open and advanced large-scale video generative models.
\newblock \emph{arXiv preprint arXiv:2503.20314}, 2025{\natexlab{a}}.

\bibitem[Wang et~al.(2024{\natexlab{a}})Wang, Tian, Zhang, Guan, Luo, Shen, Jiang, Gu, Han, and Yang]{wang2024vexpress}
Cong Wang, Kuan Tian, Jun Zhang, Yonghang Guan, Feng Luo, Fei Shen, Zhiwei Jiang, Qing Gu, Xiao Han, and Wei Yang.
\newblock V-express: Conditional dropout for progressive training of portrait video generation.
\newblock \emph{arXiv preprint arXiv:2406.02511}, 2024{\natexlab{a}}.

\bibitem[Wang et~al.(2023)Wang, Yuan, Chen, Zhang, Wang, and Zhang]{wang2023modelscope}
Jiuniu Wang, Hangjie Yuan, Dayou Chen, Yingya Zhang, Xiang Wang, and Shiwei Zhang.
\newblock Modelscope text-to-video technical report.
\newblock \emph{arXiv preprint arXiv:2308.06571}, 2023.

\bibitem[Wang et~al.(2024{\natexlab{b}})Wang, Bai, Tan, Wang, Fan, Bai, Chen, Liu, Wang, Ge, et~al.]{wang2024qwen2}
Peng Wang, Shuai Bai, Sinan Tan, Shijie Wang, Zhihao Fan, Jinze Bai, Keqin Chen, Xuejing Liu, Jialin Wang, Wenbin Ge, et~al.
\newblock Qwen2-vl: Enhancing vision-language model's perception of the world at any resolution.
\newblock \emph{arXiv preprint arXiv:2409.12191}, 2024{\natexlab{b}}.

\bibitem[Wang et~al.(2021)Wang, Mallya, and Liu]{wang2021facev2v}
Ting-Chun Wang, Arun Mallya, and Ming-Yu Liu.
\newblock One-shot free-view neural talking-head synthesis for video conferencing.
\newblock In \emph{Proceedings of the IEEE/CVF conference on computer vision and pattern recognition}, pp.\  10039--10049, 2021.

\bibitem[Wang et~al.(2020)Wang, Bilinski, Bremond, and Dantcheva]{cvideogan}
Yaohui Wang, Piotr Bilinski, Francois Bremond, and Antitza Dantcheva.
\newblock Imaginator: Conditional spatio-temporal gan for video generation.
\newblock In \emph{Proceedings of the IEEE/CVF Winter Conference on Applications of Computer Vision}, pp.\  1160--1169, 2020.

\bibitem[Wang et~al.(2022)Wang, Li, Li, He, Huang, Zhao, Zhang, Xu, Liu, Wang, et~al.]{wang2022internvideo}
Yi~Wang, Kunchang Li, Yizhuo Li, Yinan He, Bingkun Huang, Zhiyu Zhao, Hongjie Zhang, Jilan Xu, Yi~Liu, Zun Wang, et~al.
\newblock Internvideo: General video foundation models via generative and discriminative learning.
\newblock \emph{arXiv preprint arXiv:2212.03191}, 2022.

\bibitem[Wang et~al.(2024{\natexlab{c}})Wang, Li, Zeng, Fang, Guo, Liu, Tan, Chen, Xue, Dai, and Lin]{DBLP:conf/nips/00010ZF0LTCX0L24}
Zhenzhi Wang, Yixuan Li, Yanhong Zeng, Youqing Fang, Yuwei Guo, Wenran Liu, Jing Tan, Kai Chen, Tianfan Xue, Bo~Dai, and Dahua Lin.
\newblock Humanvid: Demystifying training data for camera-controllable human image animation.
\newblock In \emph{Advances in Neural Information Processing Systems}, 2024{\natexlab{c}}.

\bibitem[Wang et~al.(2025{\natexlab{b}})Wang, Li, Zeng, Guo, Lin, Xue, and Dai]{wang2025multi}
Zhenzhi Wang, Yixuan Li, Yanhong Zeng, Yuwei Guo, Dahua Lin, Tianfan Xue, and Bo~Dai.
\newblock Multi-identity human image animation with structural video diffusion.
\newblock \emph{arXiv preprint arXiv:2504.04126}, 2025{\natexlab{b}}.

\bibitem[Wu et~al.(2023{\natexlab{a}})Wu, Zhang, Zhang, Chen, Liao, Li, Gao, Wang, Zhang, Sun, et~al.]{wu2023q}
Haoning Wu, Zicheng Zhang, Weixia Zhang, Chaofeng Chen, Liang Liao, Chunyi Li, Yixuan Gao, Annan Wang, Erli Zhang, Wenxiu Sun, et~al.
\newblock Q-align: Teaching lmms for visual scoring via discrete text-defined levels.
\newblock \emph{arXiv preprint arXiv:2312.17090}, 2023{\natexlab{a}}.

\bibitem[Wu et~al.(2023{\natexlab{b}})Wu, Ge, Wang, Lei, Gu, Shi, Hsu, Shan, Qie, and Shou]{wu2023tune}
Jay~Zhangjie Wu, Yixiao Ge, Xintao Wang, Stan~Weixian Lei, Yuchao Gu, Yufei Shi, Wynne Hsu, Ying Shan, Xiaohu Qie, and Mike~Zheng Shou.
\newblock Tune-a-video: One-shot tuning of image diffusion models for text-to-video generation.
\newblock In \emph{Proceedings of the IEEE/CVF International Conference on Computer Vision}, pp.\  7623--7633, 2023{\natexlab{b}}.

\bibitem[Xiao et~al.(2024)Xiao, Wu, Xu, Dai, Hu, Lu, Zeng, Liu, and Yuan]{xiao2024florence}
Bin Xiao, Haiping Wu, Weijian Xu, Xiyang Dai, Houdong Hu, Yumao Lu, Michael Zeng, Ce~Liu, and Lu~Yuan.
\newblock Florence-2: Advancing a unified representation for a variety of vision tasks.
\newblock In \emph{Proceedings of the IEEE/CVF Conference on Computer Vision and Pattern Recognition}, pp.\  4818--4829, 2024.

\bibitem[Xie et~al.(2022)Xie, Wang, Zhang, Dong, and Shan]{xie2022vfhq}
Liangbin Xie, Xintao Wang, Honglun Zhang, Chao Dong, and Ying Shan.
\newblock Vfhq: A high-quality dataset and benchmark for video face super-resolution.
\newblock In \emph{Proceedings of the IEEE/CVF Conference on Computer Vision and Pattern Recognition}, pp.\  657--666, 2022.

\bibitem[Xu et~al.(2024{\natexlab{a}})Xu, Li, Su, Shang, Zhang, Liu, Wang, Van~Gool, Yao, and Zhu]{xu2024hallo}
Mingwang Xu, Hui Li, Qingkun Su, Hanlin Shang, Liwei Zhang, Ce~Liu, Jingdong Wang, Luc Van~Gool, Yao Yao, and Siyu Zhu.
\newblock Hallo: Hierarchical audio-driven visual synthesis for portrait image animation.
\newblock \emph{arXiv preprint arXiv:2406.08801}, 2024{\natexlab{a}}.

\bibitem[Xu et~al.(2024{\natexlab{b}})Xu, Chen, Guo, Yang, Li, Zang, Zhang, Tong, and Guo]{xu2024vasa}
Sicheng Xu, Guojun Chen, Yu-Xiao Guo, Jiaolong Yang, Chong Li, Zhenyu Zang, Yizhong Zhang, Xin Tong, and Baining Guo.
\newblock Vasa-1: Lifelike audio-driven talking faces generated in real time.
\newblock \emph{arXiv preprint arXiv:2404.10667}, 2024{\natexlab{b}}.

\bibitem[Yang et~al.(2023)Yang, Zeng, Yuan, and Li]{dwpose}
Zhendong Yang, Ailing Zeng, Chun Yuan, and Yu~Li.
\newblock Effective whole-body pose estimation with two-stages distillation.
\newblock In \emph{Proceedings of the IEEE/CVF International Conference on Computer Vision}, pp.\  4210--4220, 2023.

\bibitem[Ye et~al.(2023)Ye, Zhang, Liu, Han, and Yang]{ye2023ip}
Hu~Ye, Jun Zhang, Sibo Liu, Xiao Han, and Wei Yang.
\newblock Ip-adapter: Text compatible image prompt adapter for text-to-image diffusion models.
\newblock \emph{arXiv preprint arXiv:2308.06721}, 2023.

\bibitem[Ye et~al.(2022)Ye, Jiang, Ren, Liu, He, and Zhao]{GeneFace}
Zhenhui Ye, Ziyue Jiang, Yi~Ren, Jinglin Liu, Jinzheng He, and Zhou Zhao.
\newblock Geneface: Generalized and high-fidelity audio-driven 3d talking face synthesis.
\newblock In \emph{The Eleventh International Conference on Learning Representations}, 2022.

\bibitem[Yu et~al.(2023)Yu, Lezama, Gundavarapu, Versari, Sohn, Minnen, Cheng, Birodkar, Gupta, Gu, et~al.]{yu20233DVAE}
Lijun Yu, Jos Lezama, Nitesh~B Gundavarapu, Luca Versari, Kihyuk Sohn, David Minnen, Yong Cheng, Vighnesh Birodkar, Agrim Gupta, Xiuye Gu, et~al.
\newblock Language model beats diffusion--tokenizer is key to visual generation.
\newblock \emph{arXiv preprint arXiv:2310.05737}, 2023.

\bibitem[Yuan et~al.(2024)Yuan, Huang, He, Ge, Shi, Chen, Luo, and Yuan]{yuan2024identity}
Shenghai Yuan, Jinfa Huang, Xianyi He, Yunyuan Ge, Yujun Shi, Liuhan Chen, Jiebo Luo, and Li~Yuan.
\newblock Identity-preserving text-to-video generation by frequency decomposition.
\newblock \emph{arXiv preprint arXiv:2411.17440}, 2024.

\bibitem[Zhang et~al.(2023)Zhang, Cun, Wang, Zhang, Shen, Guo, Shan, and Wang]{zhang2023sadtalker}
Wenxuan Zhang, Xiaodong Cun, Xuan Wang, Yong Zhang, Xi~Shen, Yu~Guo, Ying Shan, and Fei Wang.
\newblock Sadtalker: Learning realistic 3d motion coefficients for stylized audio-driven single image talking face animation.
\newblock In \emph{Proceedings of the IEEE/CVF Conference on Computer Vision and Pattern Recognition}, pp.\  8652--8661, 2023.

\bibitem[Zhang et~al.(2024)Zhang, Gu, Wang, Wang, Cheng, Zhu, and Zou]{zhang2024mimicmotion}
Yuang Zhang, Jiaxi Gu, Li-Wen Wang, Han Wang, Junqi Cheng, Yuefeng Zhu, and Fangyuan Zou.
\newblock Mimicmotion: High-quality human motion video generation with confidence-aware pose guidance.
\newblock \emph{arXiv preprint arXiv:2406.19680}, 2024.

\bibitem[Zhao \& Zhang(2022)Zhao and Zhang]{zhao2022tps}
Jian Zhao and Hui Zhang.
\newblock Thin-plate spline motion model for image animation.
\newblock In \emph{Proceedings of the IEEE/CVF Conference on Computer Vision and Pattern Recognition}, pp.\  3657--3666, 2022.

\bibitem[Zhou et~al.(2022)Zhou, Wang, Yan, Lv, Zhu, and Feng]{zhou2022magicvideo}
Daquan Zhou, Weimin Wang, Hanshu Yan, Weiwei Lv, Yizhe Zhu, and Jiashi Feng.
\newblock Magicvideo: Efficient video generation with latent diffusion models.
\newblock \emph{arXiv preprint arXiv:2211.11018}, 2022.

\bibitem[Zhu et~al.(2022)Zhu, Wu, Zhu, Jiang, Tang, Zhang, Liu, and Loy]{zhu2022celebv}
Hao Zhu, Wayne Wu, Wentao Zhu, Liming Jiang, Siwei Tang, Li~Zhang, Ziwei Liu, and Chen~Change Loy.
\newblock Celebv-hq: A large-scale video facial attributes dataset.
\newblock In \emph{European conference on computer vision}, pp.\  650--667. Springer, 2022.

\bibitem[Zhu et~al.(2023)Zhu, Liu, Liu, Qian, Liu, and Yu]{zhu2023taming}
Lingting Zhu, Xian Liu, Xuanyu Liu, Rui Qian, Ziwei Liu, and Lequan Yu.
\newblock Taming diffusion models for audio-driven co-speech gesture generation.
\newblock In \emph{Proceedings of the IEEE/CVF Conference on Computer Vision and Pattern Recognition}, pp.\  10544--10553, 2023.

\end{thebibliography}
